\documentclass[conference]{IEEEtran}
\usepackage{times}

\usepackage{amsfonts}
\usepackage{amsopn}
\usepackage{amsmath}
\usepackage{amssymb}
\usepackage{mathtools}
\usepackage{bbm}
\usepackage{color}
\usepackage{placeins}
\usepackage{wrapfig}
\usepackage{float}
\usepackage{subcaption}
\usepackage{nicefrac}

\makeatletter  % Check IEEE style file for the reason of this. Makes hyperrefs to bib items work.
\let\NAT@parse\undefined
\makeatother

\usepackage[bookmarks=true]{hyperref}
\usepackage[ruled, linesnumbered]{algorithm2e}  %% ICML template does not liek 2e version.

\newcommand{\citep}{\cite}

\newcommand{\E}[2]{\operatorname{\mathbb{E}}_{#1}\left[#2\right]}
\newcommand{\Ebegin}[2]{\operatorname{\mathbb{E}}_{#1}\bigg[#2\bigg.}
\newcommand{\Eend}[1]{\bigg.#1\bigg]}

\newcommand{\kl}[2]{\mathrm{D_{KL}}\left(#1\;\middle\|\;#2\right)}
\newcommand{\entropy}{\mathcal{H}}
\newcommand{\policy}{\pi}
\newcommand{\ent}{\mathcal{H}}

\newcommand{\voidarg}{{\,\cdot\,}}

\newcommand{\sspace}{\mathcal{S}}
\newcommand{\aspace}{\mathcal{A}}

\newcommand{\state}{\mathbf{s}}
\newcommand{\sz}{{\state_0}}

\newcommand{\st}{{\state_t}}

\newcommand{\sT}{{\state_T}}
\newcommand{\stp}{{\state_{t+1}}}

\newcommand{\action}{\mathbf{a}}

\newcommand{\at}{{\action_t}}

\newcommand{\aT}{{\action_T}}

\newcommand{\opt}{^*}

\newcommand{\qparams}{\theta}
\newcommand{\pparams}{\phi}

\newcommand{\aref}[1]{\hyperref[#1]{Appendix~\ref*{#1}}}

\def\alignautorefname~#1\null{%
  (#1)\null
}\def\equationautorefname~#1\null{%
  Equation~#1\null
}
\usepackage[utf8]{inputenc} %
\usepackage[T1]{fontenc}    %
\usepackage{hyperref}       %
\usepackage{url}            %
\usepackage{booktabs}       %
\usepackage{amsfonts}       %
\usepackage{nicefrac}       %
\usepackage{microtype}      %
\usepackage{flushend}
\usepackage[numbers,sort&compress]{natbib}
\usepackage{multicol}

\title{Learning to Walk via Deep Reinforcement Learning}

\author{
\authorblockN{Tuomas Haarnoja$^{*,1,2}$, Sehoon Ha$^{*,1}$, Aurick Zhou$^{2}$, Jie Tan$^{1}$, George Tucker$^{1}$ and Sergey Levine$^{1,2}$}
\authorblockA{$^{1}$Google Brain $^{2}$Berkeley Artificial Intelligence Research, University of California, Berkeley \\
Email: \mbox{\{tuomash,sehoonha,jietan,gjt\}@google.com},\{azhou42,svlevine\}@berkeley.edu}
$^{*}$The first two authors contributed equally.
}

\begin{document}

\maketitle
\thispagestyle{empty}
\pagestyle{empty}

\begin{abstract}
Deep reinforcement learning (deep RL) holds the promise of automating the acquisition of complex controllers that can map sensory inputs directly to low-level actions. In the domain of robotic locomotion, deep RL could enable learning locomotion skills with minimal engineering and without an explicit model of the robot dynamics. Unfortunately, applying deep RL to real-world robotic tasks is exceptionally difficult, primarily due to poor sample complexity and sensitivity to hyperparameters. While hyperparameters can be easily tuned in simulated domains, tuning may be prohibitively expensive on physical systems, such as legged robots, that can be damaged through extensive trial-and-error learning. In this paper, we propose a sample-efficient deep RL algorithm based on maximum entropy RL that requires minimal per-task tuning and only a modest number of trials to learn neural network policies.
We apply this method to learning walking gaits on a real-world Minitaur robot. Our method can acquire a stable gait from scratch directly in the real world in about two hours, without relying on any model or simulation, and the resulting policy is robust to moderate variations in the environment. We further show that our algorithm achieves state-of-the-art performance on simulated benchmarks with a single set of hyperparameters. Videos of training and the learned policy can be found on the project website\footnote[3]{\href{https://sites.google.com/view/minitaur-locomotion/}{https://sites.google.com/view/minitaur-locomotion/}}.
\end{abstract}

\section{Introduction}
Designing locomotion controllers for legged robots is a long-standing research challenge. Current state-of-the-art methods typically employ a pipelined approach, consisting of components such as state estimation, contact scheduling, trajectory optimization, foot placement planning, model-predictive control, and operational space control~\cite{gehring2016practice, hutter2016anymal, BledtPKCWK18, Apgar2018FastOT}. Designing these components requires expertise and often an accurate dynamics model of the robot that can be difficult to acquire. In contrast, end-to-end deep reinforcement learning (deep RL) does not assume any prior knowledge of the gait or the robot's dynamics, and can in principle be applied to robotic systems without explicit system identification or manual engineering. If successfully applied, deep RL can automate the controller design, completely removing the need for system identification, and resulting in gaits that are directly optimized for a particular robot and environment. However, applying deep RL to learning gaits in the real world is challenging, since current algorithms often require a large number of samples---on the order of tens of thousands of trials~\cite{tan2018sim}. Moreover, such algorithms are often highly sensitive to hyperparameter settings and require considerable tuning~\cite{henderson2017deep}, further increasing the overall sample complexity. For this reason, many prior methods have studied learning of locomotion gaits in simulation \cite{heess2017emergence, xie2018feedback, peng2016terrain, berseth2018progressive}, requiring accurate system identification and modeling.

\begin{figure}[tb]
    \centering
    \begin{subfigure}{0.45\columnwidth}
        \includegraphics[width=\textwidth, trim={20mm 0 20mm 0}, clip]{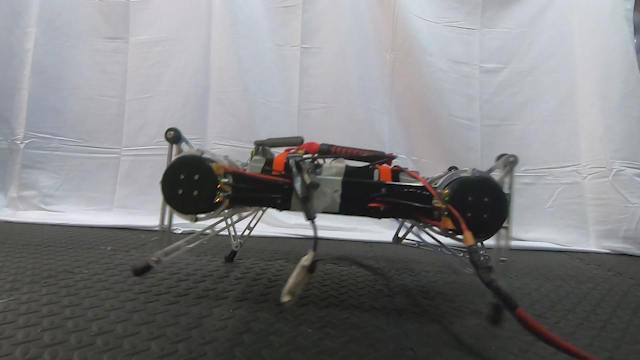}
    \end{subfigure}
    \begin{subfigure}{0.45\columnwidth}
        \includegraphics[width=\textwidth, trim={10mm 0 30mm 0}, clip]{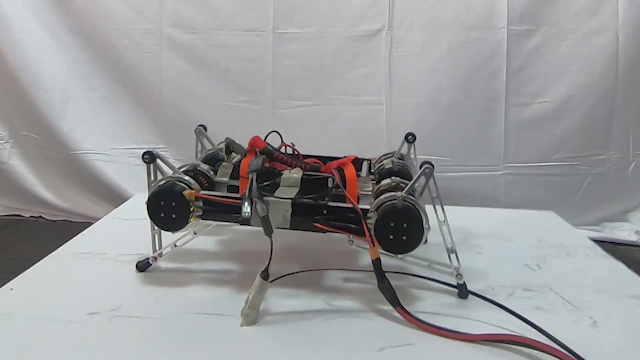}
    \end{subfigure}\\\vspace{1mm}
    \begin{subfigure}{0.45\columnwidth}
        \includegraphics[width=\textwidth, trim={20mm 0 20mm 0}, clip]{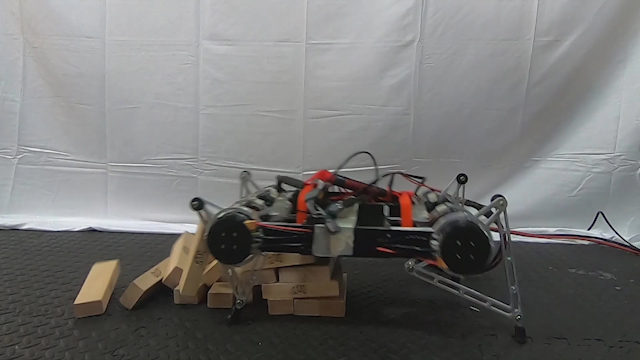}
    \end{subfigure}
    \begin{subfigure}{0.45\columnwidth}
        \includegraphics[width=\textwidth, trim={10mm 0 30mm 0}, clip]{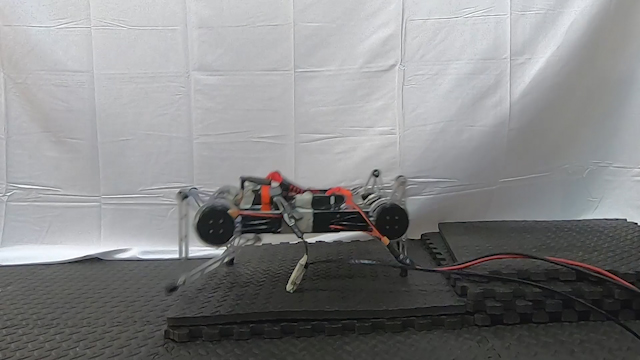}
    \end{subfigure}
    \caption{\small Illustration of a walking gait learned in the real world. The policy is trained only on a flat terrain, but the learned gait is robust and can handle obstacles that were not seen during training.}
    \label{fig:minitaur_walk}
\end{figure}

In this paper, we aim to address these challenges by developing a deep RL algorithm that is both sample efficient and robust to the choice of hyperparameters, thus allowing us to learn locomotion gaits directly in the real world, without prior modeling.
In particular, we extend the framework of maximum entropy RL. Methods of this type, such as soft actor-critic~\cite{haarnoja2018soft} and soft Q-learning~\cite{haarnoja2017reinforcement}, can achieve state-of-the-art sample efficiency~\cite{haarnoja2018soft} and have been successfully deployed in real-world manipulation tasks~\cite{haarnoja2018composable,mahmood2018benchmarking}, where they exhibit a high degree of robustness due to entropy maximization~\cite{haarnoja2018composable}. However, maximum entropy RL algorithms are sensitive to the choice of the temperature parameter, which determines the trade-off between exploration (maximizing the entropy) and exploitation (maximizing the reward). In practice, this temperature is considered as a hyperparameter that must be tuned manually for each task.

We propose an extension to the soft actor-critic algorithm~\cite{haarnoja2018soft} that removes the need for manually tuning of the temperature parameter. Our method employs gradient-based optimization of the temperature towards the targeted expected entropy over the visited states. In contrast to standard RL, our method controls only the expected entropy over the states, while the per-state entropy can still vary---a desirable property that allows the policy to automatically reduce entropy for states where acting deterministically is preferred, while still acting stochastically in other states. Consequently, our approach virtually eliminates the need for per-task hyperparameter tuning, making it practical for us to apply this algorithm to learn quadrupedal locomotion gaits directly on a real-world robotic system.

The principal contribution of our paper is an end-to-end RL framework for legged locomotion on physical robots, which includes a data efficient learning algorithm based on maximum entropy RL and an asynchronous learning system. We demonstrate the framework by training a Minitaur robot \cite{kenneally2016design} (Figure~\ref{fig:minitaur_walk}) to walk. While we train the robot on flat terrain, the learned policy can generalize to unseen terrains and is moderately robust to perturbations. The training requires about 400 rollouts, equating to about two hours of real-world time. In addition to the robot experiments, we evaluate our algorithm on simulated benchmark tasks and show that it can achieve state-of-the-art performance and, unlike prior works based on maximum entropy RL, can use exactly the same hyperparameters for all tasks.

\section{Related Work}

Current state-of-the-art locomotion controllers typically adopt a pipelined control scheme. For example, the MIT Cheetah \cite{BledtPKCWK18} uses a state machine over contact conditions, generates simple reference trajectories, performs model predictive control \cite{Carlo2018} to plan for desired contact forces, and then uses Jacobian transpose control to realize them. The ANYmal robot \cite{hutter2016anymal} plans footholds based on the inverted pendulum model \cite{raibert1986legged}, applies CMA-ES \cite{hansen2006cma} to optimize a parameterized controller \cite{gehring2014towards, gehring2016practice}, and solves a hierarchical operational space control problem \cite{hutter2014quadrupedal} to produce joint torques, contact forces, and body motion. While these methods can provide effective gaits, they require considerable prior knowledge of the locomotion task and, more importantly, of the robot's dynamics. In contrast, our method aims to control the robot without prior knowledge of either the gait or the dynamics. We do not assume access to any trajectory design, foothold planner, or a dynamics model of the robot, since all learning is done entirely through real-world interaction. The only requirement is knowledge of the dimension and bounds of the state and action space, which in our implementation correspond to joint angles, IMU readings, and desired motor positions. While in practice, access to additional prior knowledge could be used to accelerate learning (see, e.g.,~\cite{iscen18a}), end-to-end methods that make minimal prior assumptions are broadly applicable, and developing such techniques will make acquisition of gaits for diverse robots in diverse conditions scalable.

Deep RL has been used extensively to learn locomotion policies in simulation \cite{heess2017emergence, xie2018feedback, peng2016terrain, berseth2018progressive} and even transfer them to real-world robots \cite{tan2018sim, Hwangboeaau5872}, but this inevitably incurs a loss of performance due to discrepancies in the simulation, and requires accurate system identification. Using such algorithms directly in the real world has proven challenging. Real-world applications typically make use of simple and inherently stable robots \cite{ha2018automated} or low-dimensional gait parameterizations \cite{kohl2004policy,calandra2016bayesian,rai2017bayesian}, or both \cite{tedrake2005learning}. In contrast, we show that we can acquire locomotion skills directly in the real world using neural-net policies.

Our algorithm is based on maximum entropy RL, which maximizes the weighted sum of the  the expected return and the policy's expected entropy. This framework has been used in many contexts, from inverse RL~\citep{ziebart2008maximum} to optimal control~\citep{todorov2008general,toussaint2009robot,rawlik2012stochastic}. One advantage of maximum entropy RL is that it produces relatively robust policies, since injection of structured noise during training causes the policy to explore the state space more broadly and improves the robustness of the policy~\cite{haarnoja2018composable}. However, the weight on the entropy term (``temperature'') is typically chosen heuristically~\cite{o2016pgq,haarnoja2017reinforcement,nachum2017bridging,schulman2017equivalence,haarnoja2018soft}. In our observation, this parameter is very sensitive and manual tuning can make real-world application of the maximum entropy framework difficult. Instead, we propose to constrain the expected entropy of the policy and adjust the temperature automatically to satisfy the constraint. Our formulation is an instance of constrained MDP, which has been studied recently in \cite{bohez2019value,achiam2017constrained,tessler2018reward}. These works consider constraints that depend on the policy only via the sampling distribution, whereas in our case the constraint depends the policy explicitly. Our approach is also closely related to KL-divergence constraints that limit the policy change between iterations \cite{peters2006policy,schulman2015trust,abdolmaleki2018maximum} but is applied directly to the current policy's entropy. We find that this simple modification drastically reduces the effort of parameter tuning on both simulated benchmarks and our robotic locomotion task.

\section{Asynchronous Learning System}
\begin{figure}[tb]
    \centering
    \includegraphics[width=0.36\textwidth]{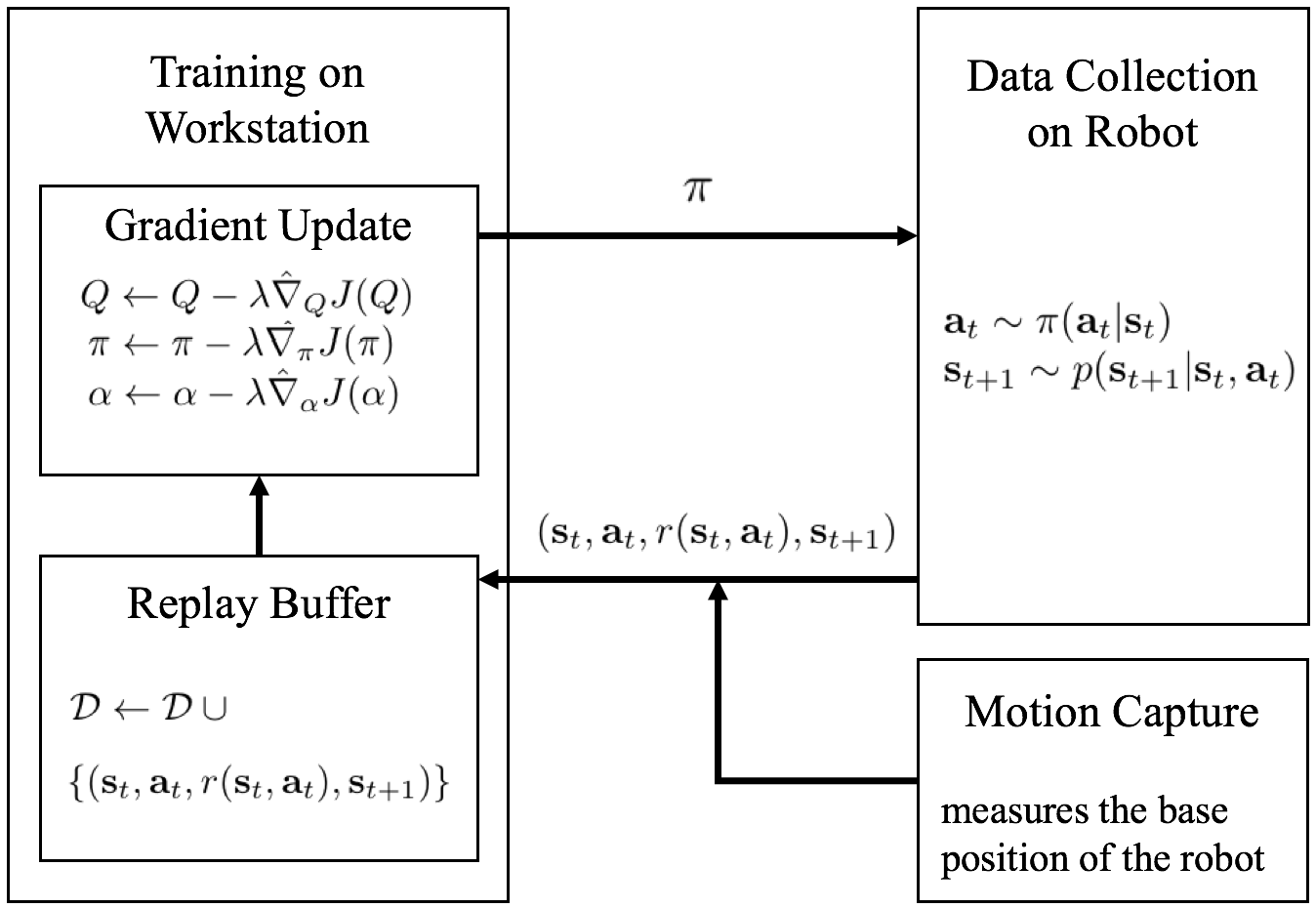}
    \caption{\small Overview of our learning system. The learning system runs the training and data collection asynchronously across multiple machines.}
	\label{fig:pipeline}
\end{figure}
In this section, we will first describe our asynchronous robotic RL system, which we will use to evaluate real-world RL for robotic locomotion. The system, shown in Figure \ref{fig:pipeline}, consists of three components: a data collection job that collects  robot experience, a motion capture job that computes the reward signal based on robot's position measured by a motion capture system, and a training job that updates the neural networks. These subsystems run asynchronously on different machines. When the learning system starts, the subsystems are synchronized to a common clock and use timestamps to sync the future data streams.

The data collection job runs on an on-board computer and executes the latest policy $\pi$ produced by the training job. For each control step $t$, it collects observations $\st$, performs neural network policy inference, and executes an action $\at$. The entire observed trajectory, or rollout, is recorded into tuples ${(\st, \at, \stp)}_{t=0,...,N-1}$ and sent to the training job. The motion capture system measures the position of the robot and provides the reward signal $r(\st,\at)$. 
It periodically pulls data from the robot and the motion capture system, evaluates the reward function, and appends it to a replay buffer. The training subsystem runs on a workstation. At each iteration of training, the training job randomly samples a batch of data from this buffer and uses stochastic gradient descent to update the value network, the policy network, and the temperature parameter, as we will discuss in \autoref{sec:algorithm}. Once training is started, minimal human intervention is needed, except for the need to reset the robot if it falls or runs out of free space.

The asynchronous design allows us to pause or restart any subsystem without affecting the other subsystems. In practice, we found this particularly useful because we often encounter hardware and communication errors, in which case we can safely restart any of the subsystems without impacting the entire learning process. In addition, our system can be easily scaled to multiple robots by simply increasing the number of data collection jobs. In the following sections, we describe our proposed reinforcement learning method in detail.

\section{Reinforcement Learning Preliminaries}
Reinforcement learning aims to learn a policy that maximizes the expected sum of rewards~\cite{sutton1998reinforcement}. We consider Markov decision processes where the state space $\mathcal{S}$ and action space $\mathcal{A}$ are continuous. An agent starts at an initial state $\state_0 \sim p(\state_0)$, samples an action $\action_t$ from a policy $\pi(\voidarg | \st)\in\Pi$, receives a bounded reward $r(\state_t, \action_t)$, and transitions to a new state $\stp$ according to the dynamics $p(\voidarg | \st,\at)$. This generates a trajectory of states and actions $\tau = (\state_0,\action_0, \state_1, \action_1, \ldots)$. We denote the trajectory distribution induced by $\policy$ by $\rho_\pi(\tau)=p(\sz)\prod_t\pi_t(\at|\st)p(\stp|\st,\at)$, and we overload the notation and use $\rho_\policy(\st, \at)$ and $\rho_\policy(\st)$ to denote the corresponding state-action and state marginals, respectively.

Maximum entropy RL optimizes both the expected return and the entropy of the policy. For finite-horizon MDPs, the corresponding objective can be expressed as
\begin{equation}
    J(\pi) = \sum_{t = 0}^{T} \E{\tau\sim \rho_\policy}{r(\st, \at) - \alpha_t \log \pi_t(\at|\st)},
    \label{eq:max_ent_objective}
\end{equation}
which incentivizes the policy to explore more widely improving its robustness against perturbations~\cite{haarnoja2018composable}. The temperature parameter $\alpha$ determines the relative importance of the entropy term against the reward, and thus controls the stochasticity of the optimal policy. The maximum entropy objective differs from the standard maximum expected reward objective used in conventional reinforcement learning, though the conventional objective can be recovered in the limit as $\alpha \rightarrow 0$. In the finite horizon case, the policy is time dependent, and we write $\pi$ and $\alpha$ to denote the set of all policies $(\pi_0,\pi_1,...,\pi_T)$ or temperatures $(\alpha_0,\alpha_1,...,\alpha_T)$. We can extend the objective to infinite horizon problems by introducing a discount factor $\gamma$ to ensure that the sum of expected rewards and entropies is finite \cite{haarnoja2017reinforcement}, in which case we overload the notation and denote a stationary policy and temperature as $\pi$ and $\alpha$.

One of the central challenges with the objective in~\eqref{eq:max_ent_objective} is that the trade-off between maximizing the return, or exploitation, versus the entropy, or exploration, is directly affected by the scale of the reward function\footnote{Reward scale is the reciprocal of temperature. We will use these two terms interchangeably throughout this paper.}. Unlike in conventional RL, where the optimal policy is independent of scaling of the reward function, in maximum entropy RL the scaling factor has to be tuned per environment, and a sub-optimal scale can drastically degrade the performance~\cite{haarnoja2018soft}. 

\section{Automating Entropy Adjustment for Maximum Entropy RL}
\label{sec:algorithm}

Learning robotic tasks in the real world requires an algorithm that is sample efficient, robust, and insensitive to the choice of the hyperparameters. Maximum entropy RL is both sample efficient and robust, making it a good candidate for real-world robot learning~\cite{haarnoja2018composable}. However, one of the major challenges of maximum entropy RL is its sensitivity to the temperature parameter, which typically needs to be tuned for each task separately. In this section, we propose an algorithm that enables automated temperature adjustment at training time, substantially reducing the effort of hyperparameter tuning and making deep RL a viable solution for real-world robotic problems.

\subsection{Entropy Constrained Objective}
The magnitude of the reward differs not only across tasks, but it also depends on the policy, which improves over time during training. Since the optimal entropy depends on this magnitude, choosing the ideal temperature is particularly difficult: the entropy can vary unpredictably both across tasks and during training as the policy becomes better. Instead of requiring the user to set the temperature manually, we can automate this process by formulating a modified RL objective, where the entropy is treated as a constraint. Simply forcing the entropy to a fixed value is a poor solution, since the policy should be free to explore more in regions where the optimal action is uncertain, but remain more deterministic in states with a clear distinction between good and bad actions. Therefore, we constrain the \emph{expected} entropy of the policy, while the entropy at different states can still vary. We show that the Lagrangian relaxation of this problem leads to the maximum entropy objective with respect to the policy, where the dual variable takes the role of the temperature.

In particular, our aim is to find a stochastic policy with maximal expected return that satisfies a minimum expected entropy constraint. Formally, we want to solve the constrained optimization problem
\begin{align}
    \max_{\pi\in\Pi}& \E{\tau\sim\rho_\pi}{\sum_{t=0}^T r(\st,\at)}\notag\\
    &\text{ s.t. } \E{(\st,\at)\sim\rho_\pi}{-\log \left(\pi_t(\at|\st)\right)} \geq \entropy,\ \forall t,
    \label{eq:constrained_rl}
\end{align}
where $\ent$ is the desired minimum expected entropy. Note that, for fully observed MDPs, the policy that optimizes the expected return is deterministic, so we expect this constraint to usually be tight and do not need to impose an upper bound on the entropy.  

We start by writing out the Lagrangian relaxation of \eqref{eq:constrained_rl}, as typical in the prior works~\cite{bohez2019value,achiam2017constrained, tessler2017deep}: 
\begin{align}
\resizebox{\columnwidth}{!}{$
    \mathcal{L}(\policy, \alpha) = \E{\tau\sim\rho_\pi}{\sum_{t=0}^T r(\st,\at) + \alpha_t \left(-\log \left(\pi_t(\at|\st)\right) - \entropy\right)}.
    \label{eq:lagrangian}$}
\end{align}
We optimize this objective using the dual gradient method.
Note that for a fixed dual variable, the Lagrangian is exactly equal to the maximum entropy objective in \eqref{eq:max_ent_objective} minus an additive constant ($\alpha_t\mathcal{H}$) per time step, and can thus be optimized with any off-the-shelf maximum entropy RL algorithm. Specifically, we resort to approximate dynamic programming, which turn out to correspond to the soft actor-critic algorithm~\cite{haarnoja2018soft}.
We first define the soft Q-function and use it to bootstrap the algorithm. The optimal soft Q-function is defined as
\begin{align}
    Q_t^*(\st, \at) = r(\st,\at) + \E{\stp\sim\rho_\policy}{V_{t+1}^*(\stp)},
    \label{eq:q_function}
\end{align}
where
\begin{align}
    V^*_t(\st) = \Ebegin{\at\sim\policy^*_{t}}{Q^*_{t}(\st,\at)}\Eend{- \alpha_{t}\log \left(\pi^*_{t}(\at|\st)\right)},%
\end{align}
and $\pi_t^*$ denotes the optimal policy for time $t$. We have omitted the dependency of the soft Q-function on the dual variable of the future time steps for brevity. We also abuse the notation slightly, and write $Q_t^*$ to denote $Q_t^{\pi^*}$, which are equal only if $\Pi$ is a set of all policies (and not for example the set of Gaussian policies). We initialize the iteration by setting $Q^*_T(\sT,\aT) = r(\sT,\aT)$.  Assuming we have evaluated $Q_t^*$ for some $t$, we can substitute it to the Lagrangian. We can now solve for the optimal policy at time $t$ for all $\st\in\sspace$ by noting that the optimal policy at time $t$ is independent of the policy at the previous time steps:
\begin{align}
  \policy\opt_t(\voidarg|\st) &\in \arg \max_{\policy_t\in\Pi} \E{\at\sim\pi_t}{Q^*_{t}(\state_{t}, \action_{t}) - \alpha_{t} \log \policy_t(\action_{t}|\state_{t})}\notag\\
&= \arg\min_{\policy_t\in\Pi} \kl{\policy_t(\voidarg|\st)}{\frac{\exp\left(\frac{1}{\alpha_t}Q_t^*(\st, \voidarg)\right)}{Z_t(\st)}}.
\label{eq:optimal_policy}
\end{align}
The partition function $Z_t(\st) = \int_{\aspace}\exp\left(\frac{1}{\alpha_t}Q\opt_t(\st, \at)\right) d\at$ does not depend on $\policy_t\opt$, so we can ignore it for optimizing $\policy_t\opt$. This is exactly the soft policy improvement step introduced by~\cite{haarnoja2018soft}, with an additional temperature parameter $\alpha_t$.  In contrast to~\cite{haarnoja2018soft}, which shows that this  update leads to an improvement in the infinite horizon case, we derived it starting from the finite horizon objective. By traversing backwards in time, we can optimize the Lagrangian with respect to the policy.

After solving for the policy for a fixed dual variable, we improve the dual in order to satisfy the entropy constraint. 
We can optimize the temperature by moving it in the direction of the negative gradient of \eqref{eq:lagrangian}:
\begin{align}
    \alpha_t \leftarrow \alpha_t + \lambda_\alpha \E{(\st,\at)\sim\rho_{\pi^*}}{\log\left(\pi^*_t(\at|\st) + \mathcal{H}\right)},
    \label{eq:dual_update}
\end{align}
where $\lambda_\alpha$ is the learning rate\footnote{We also need to make sure  $\alpha_t$ remains non-negative. In practice, we thus parameterize $\alpha_t = \exp\left(\beta_t \right)$ and optimize $\beta_t$ instead.}. The equations \eqref{eq:q_function}, \eqref{eq:optimal_policy}, and \eqref{eq:dual_update} constitute the core of our algorithm. However, solving these equations exactly is not practical for continuous state and actions, and in practice, we cannot compute the expectations, but instead have access to unbiased samples. Therefore, for a practical algorithm, we need to resort to function approximators and stochastic gradient descent as well as other standard tricks to stabilize training, as discussed in the next section.

\subsection{Practical Algorithm}     
In practice, we parameterize a Gaussian policy with parameters $\pparams$, and learn them using stochastic gradient descent for the discounted, infinite horizon problem. We additionally use two parameterized Q-functions, with parameters $\qparams_1$ and $\qparams_2$, as suggested in~\cite{haarnoja2018soft}. We learn the Q-function parameters as a regression problem by minimizing the following loss $J_Q(\qparams_i)$:
\begin{align}
\E{(\st, \at, \stp) \sim  \mathcal{D}}{\left(Q_{\qparams_i}(\st, \at) - (r(\st, \at) + \gamma V_{\qparams_1, \qparams_2}(\stp)) \right)^2}
\label{eq:q_objective}
\end{align}
using minibatches from a replay buffer $\mathcal{D}$. The value function $V_{\qparams_1, \qparams_2}(\st)$ is implicitly defined through the Q-functions and the policy as $\E{\at\sim\policy_\pparams}{\underset{i\in\{1,2\}}{\min} Q_{\qparams_i}(\st, \at) - \alpha \log \policy_\pparams(\at|\st)}$. We learn a Gaussian policy by minimizing 
\begin{align}
    J_\policy(\pparams) = \E{\st\sim\mathcal{D}, \at\sim\policy_\pparams}{\alpha \log \policy_\pparams(\at|\st) - \min_{i\in\{1,2\}}Q_{\qparams_i}(\st, \at)},
\label{eq:policy_objective}
\end{align}
using the reparameterization trick~\cite{kingma2013auto}. This procedure is the same as the standard soft actor-critic algorithm~\cite{haarnoja2018soft}, but with an explicit, dynamic temperature $\alpha$.

To learn $\alpha$, we need to minimize the dual objective, which can be done by approximating dual gradient descent. %
Instead of optimizing with respect to the primal variables to convergence, we use a truncated version that performs incomplete optimization and alternates between taking a single gradient step on each objective. %
While convergence to the global optimum is not guaranteed,
we found this approach to work well in practice. Thus, we compute gradients for $\alpha$ with the following objective:
\begin{align}
J(\alpha)  = \E{\st\sim \mathcal{D}, \at\sim \policy_{\pparams}}{ - \alpha\log\pi_\pparams(\at|\st) - \alpha \ent}.
\label{eq:alpha_objective}
\end{align}
The proposed algorithm alternates between a data collection phase and an optimization phase. In the optimization phase, the algorithm optimizes all objectives in~\eqref{eq:q_objective} -- \eqref{eq:alpha_objective} jointly. We also incorporate delayed target Q-function networks as is standard in prior work. \autoref{alg:sac} summarizes the full algorithm, where $\hat \nabla$ denotes stochastic gradients.

\begin{algorithm}
\caption{Soft Actor-Critic with Automatic Entropy Adjustment}
\label{alg:sac}
Initialize function approximators parameters $\qparams_1$ $\qparams_2$, $\pparams$, and a global temperature coefficient $\alpha$.\\
\For{each iteration}{
	\For{each environment step}{
	$\at \sim \policy(\at|\st)$\\
	$\stp \sim p(\stp| \st, \at)$\\
	$\mathcal{D} \leftarrow \mathcal{D} \cup \left\{(\st, \at, r(\st, \at), \stp)\right\}.$\\
	}
	 \For{each gradient step}{
	    $\qparams_i \leftarrow \qparams_i - \lambda \hat \nabla_\qparams J_Q(\qparams_i)$ for $i\in\{1,2\}$\\
	    $\pparams \leftarrow \pparams - \lambda \hat \nabla_\pparams J_\policy(\pparams)$\\
	    $\alpha \leftarrow \alpha - \lambda \hat \nabla_\alpha J(\alpha)$\\
	 }
	 $\bar\qparams_i\leftarrow \tau \qparams_i + (1-\tau)\bar\qparams_i$ for $i\in\{1,2\}$
}
\end{algorithm}%

\section{Evaluation on Simulation Environments}
Before evaluating on real-world locomotion, we conduct a comprehensive evaluation in simulation to validate our algorithm. Our goal is to answer following four questions:
\begin{enumerate}
    \item Does our method achieve the state-of-the-art data efficiency?
    \item How sensitive is our method to the hyperparameter?
    \item Is our method effectively regulating the entropy and dynamically adjusting the temperature during learning?
    \item Can the learned policy generalize to unseen situations?
\end{enumerate}

\subsection{Evaluation on OpenAI Benchmark Environments}
\begin{figure}[t]
    \centering
	\begin{subfigure}[b]{0.235\textwidth}
        \includegraphics[width=\textwidth]{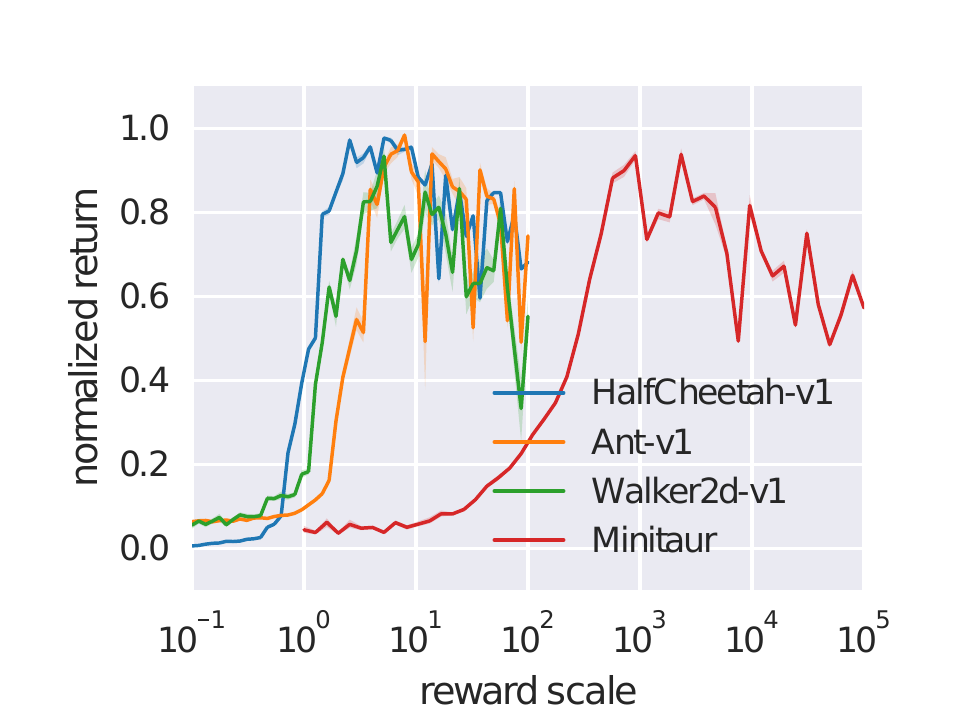}
        \caption{\footnotesize{Standard SAC over reward scale}}
	\label{fig:reward_scale_sweep}
    \end{subfigure}
	\begin{subfigure}[b]{0.235\textwidth}
        \includegraphics[width=\textwidth]{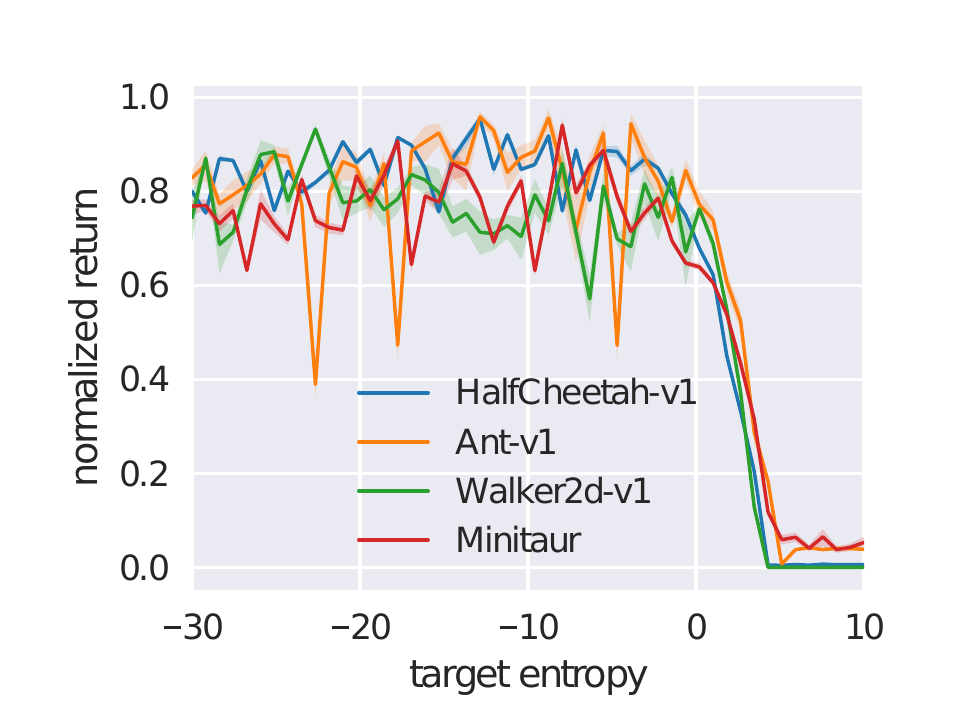}
        \caption{\footnotesize{Our method over target entropy}}
	\label{fig:target_entropy_sweep}
    \end{subfigure}
    \caption{\small Average normalized performance over the last 100k samples on a range of environments. (a) Performance of standard SAC as a function of reward scale, and (b) Performance of our method as a function of target entropy. Our method is substantially less sensitive to the choice of the hyperparameter.}
	\label{fig:hyperparameter_sweeps}
	\vspace{-0.2in}
\end{figure}

We first evaluate our algorithm on four standard benchmark environments for continuous locomotion tasks in OpenAI Gym benchmark suite~\cite{brockman2016openai}.
We compare our method to soft actor-critic (SAC)~\cite{haarnoja2018soft} with a fixed temperature parameter that is tuned for each environment. We also compare to deep deterministic policy gradient (DDPG)~\cite{lillicrap2015continuous}, proximal policy optimization (PPO)~\cite{schulman2017proximal}, and twin delayed deep deterministic policy gradient algorithm (TD3)~\cite{fujimoto2018addressing}. All of the algorithms use the same network architecture: all of the function approximators (policy and Q-functions for SAC) are parameterized with a two-layer neural network with 256 hidden units on each layer, and we use ADAM~\cite{kingma2014adam} with the same learning rate of 0.0003 to train all the networks and temperature parameter $\alpha$. For standard SAC, we tune the reward scale per environment using grid search. Poorly chosen reward scales can degrade performance drastically (see \autoref{fig:reward_scale_sweep}). For our method, we simply set the target entropy to be -1 per action dimension (i.e., HalfCheetah has target entropy -6, while Humanoid uses~-17).

\begin{figure}[tb]
    \centering
	\begin{subfigure}[b]{0.23\textwidth}
        \includegraphics[width=\textwidth]{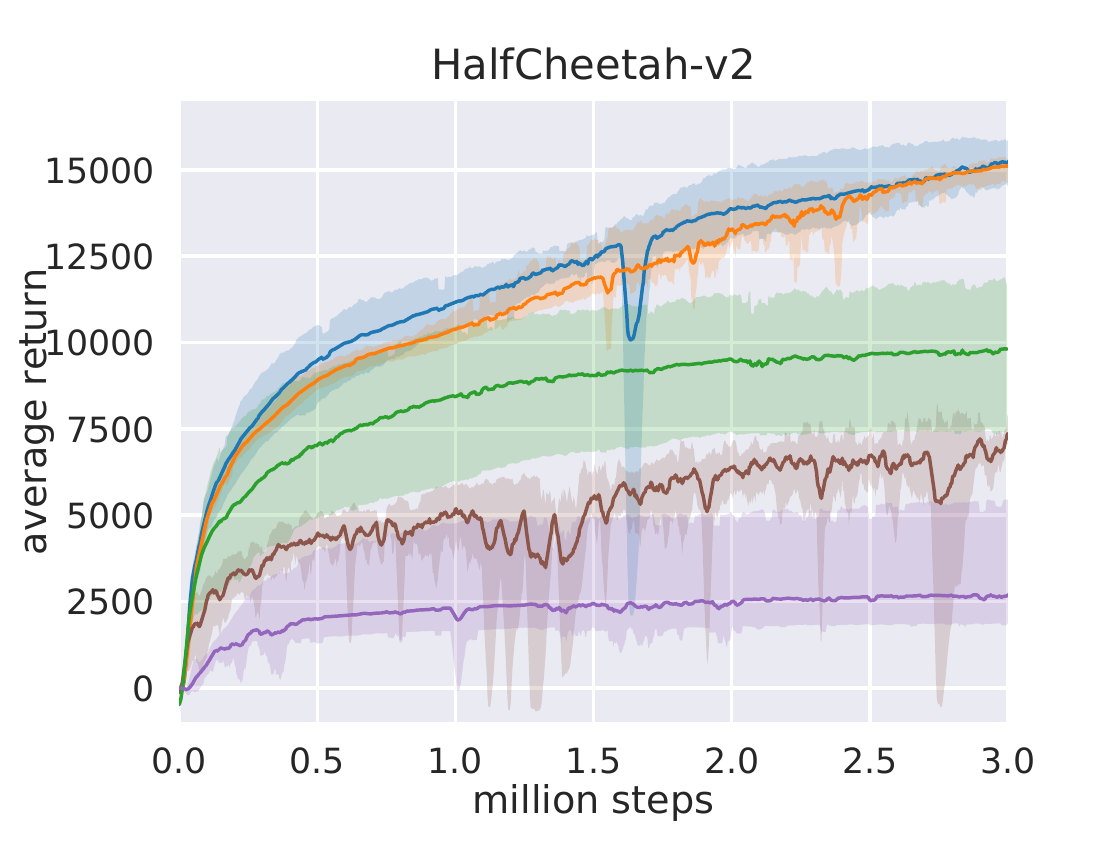}
        \caption{HalfCheetah}
    \end{subfigure}
	\begin{subfigure}[b]{0.23\textwidth}
        \includegraphics[width=\textwidth]{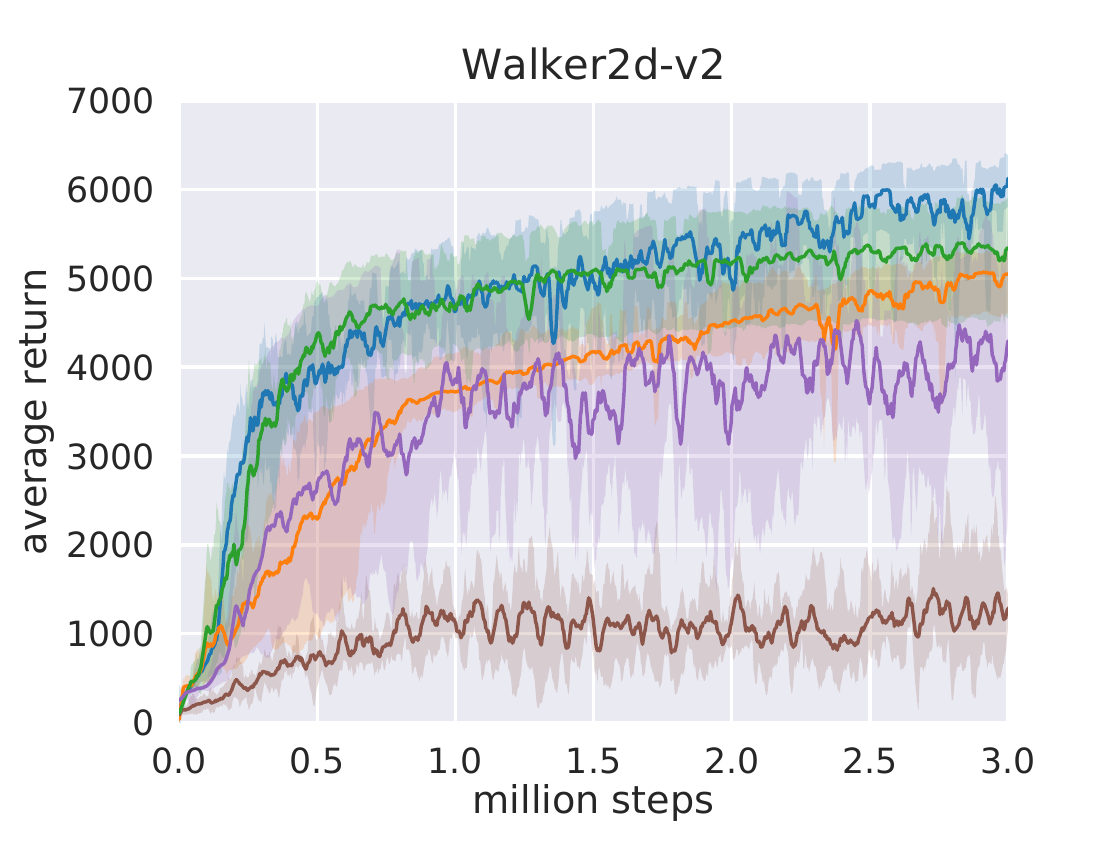}
        \caption{Walker2d}
    \end{subfigure}
	\begin{subfigure}[b]{0.23\textwidth}
        \includegraphics[width=\textwidth]{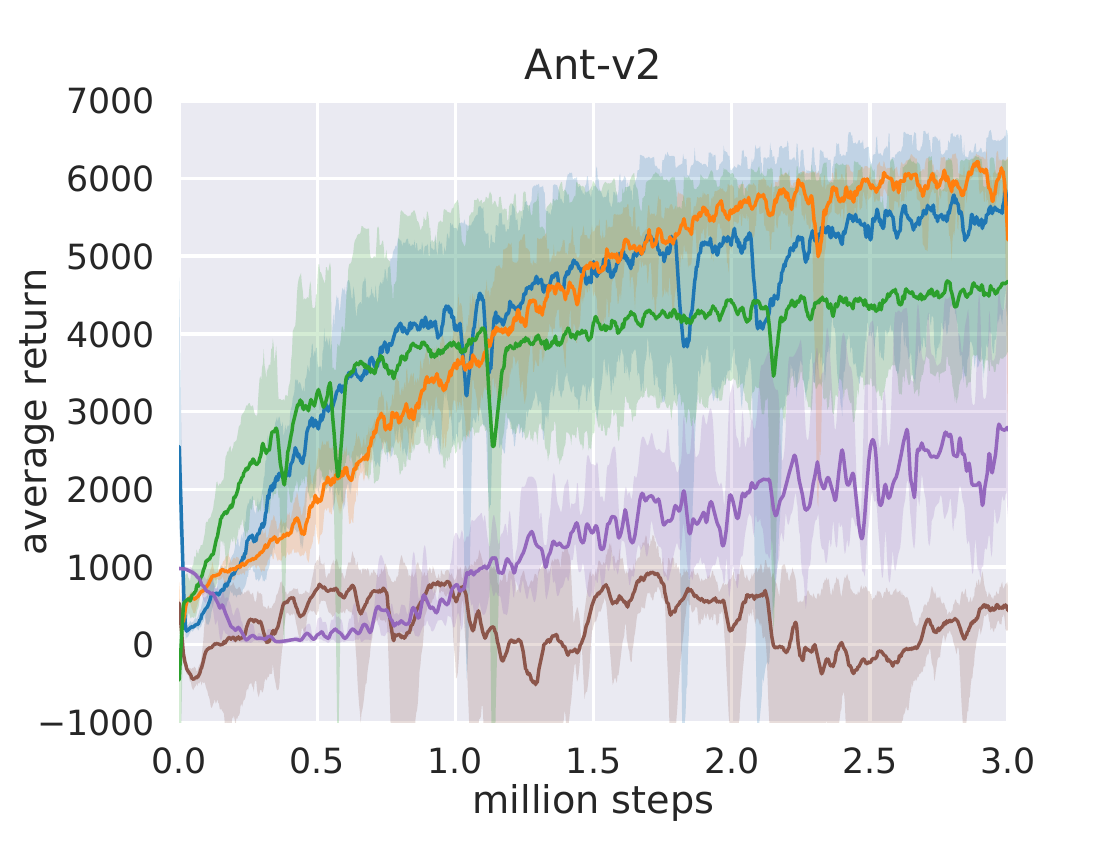}
        \caption{Ant}
    \end{subfigure}
	\begin{subfigure}[b]{0.23\textwidth}
        \includegraphics[width=\textwidth]{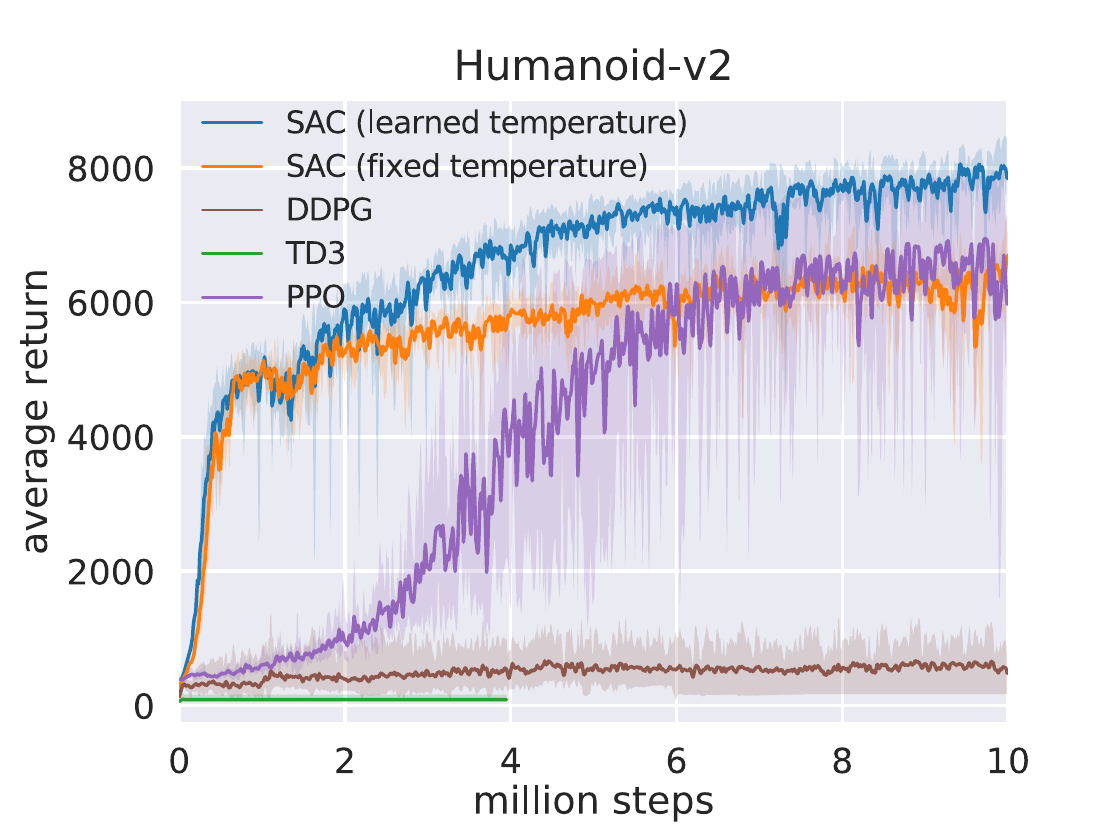}
        \caption{Humanoid}
    \end{subfigure}
    \caption{\small (a) -- (d) Standard benchmark training results. Our method (blue) achieves similar or better performance compared to other algorithms. Note that all other algorithms except ours went through dense hyperparameter tuning to achieve the above learning curves.} 
	\label{fig:benchmarks}
	\vspace{-0.2in}
\end{figure}

\subsubsection{Comparative Evaluation}
\autoref{fig:benchmarks} shows a comparison of the algorithms. The solid line denotes the average performance over five random seeds, and the shaded region corresponds to the best and worst performing seeds. The results indicate that our method (blue) achieves practically identical or better performance compared to standard SAC (orange), which is tuned per environment for all environments. Overall, our method performs better or comparably to the other baselines, standard SAC, DDPG, TD3, and PPO.

\subsubsection{Sensitivity Analysis}

We compare the sensitivity to the hyperparameter between our method (target entropy) and the standard SAC (reward scale). Both maximum entropy RL algorithms~\cite{haarnoja2018soft} and standard RL algorithms~\cite{henderson2017deep} can be very sensitive to the scale of the reward function. In the case of maximum entropy RL, this scale directly affects the trade-off between reward maximization and entropy maximization~\cite{haarnoja2018soft}. 
We first validate the sensitivity of standard SAC by running experiments on the HalfCheetah, Walker, Ant, and the simulated Minitaur robot (See Section~\ref{sec:sim_minitaur_env} for more details). 
\autoref{fig:reward_scale_sweep} shows the returns for a range of reward scale values that are normalized to the maximum reward of the given task. All benchmark environments achieve good performance for about the same range of values, between 1 to 10. On the other hand, the simulated Minitaur requires roughly two orders of magnitude larger reward scale to work properly. This result indicates that, while standard benchmarks offer high variability in terms of task dimensionality, they are homogeneous in terms of other characteristics, and testing only on the benchmarks might not generalize well to seemingly similar tasks designed for different purposes. This suggests that the good performance of our method, with the same hyperparameters, on both the benchmark tasks and the Minitaur task accurately reflects its generality and robustness. \autoref{fig:target_entropy_sweep} compares the sensitivity of our method to the target entropy on the same tasks. In this case, the range of good target entropy values is essentially the same for all environments, making hyperparameter tuning substantially less laborious. It is also worth noting that this large range indicates that our algorithm is relatively insensitive to the choice of this hyperparameter.

\begin{figure}[t]
    \centering
	\begin{subfigure}[b]{0.235\textwidth}
        \includegraphics[width=\textwidth]{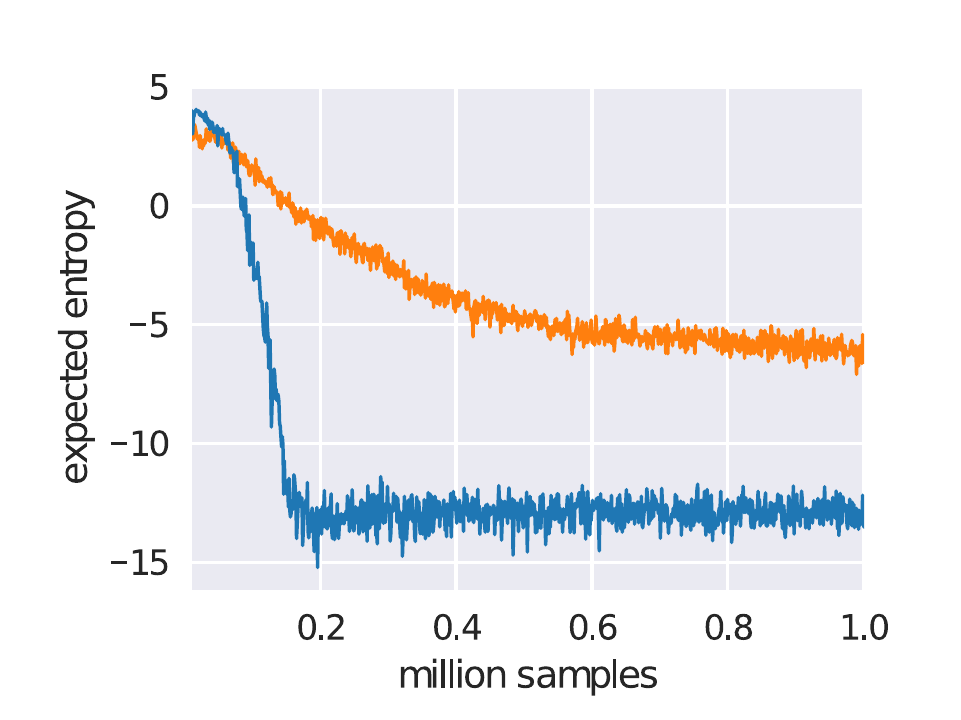}
        \caption{Entropy}
        \label{fig:entropy-half-cheetah}
    \end{subfigure}
	\begin{subfigure}[b]{0.235\textwidth}
        \includegraphics[width=\textwidth]{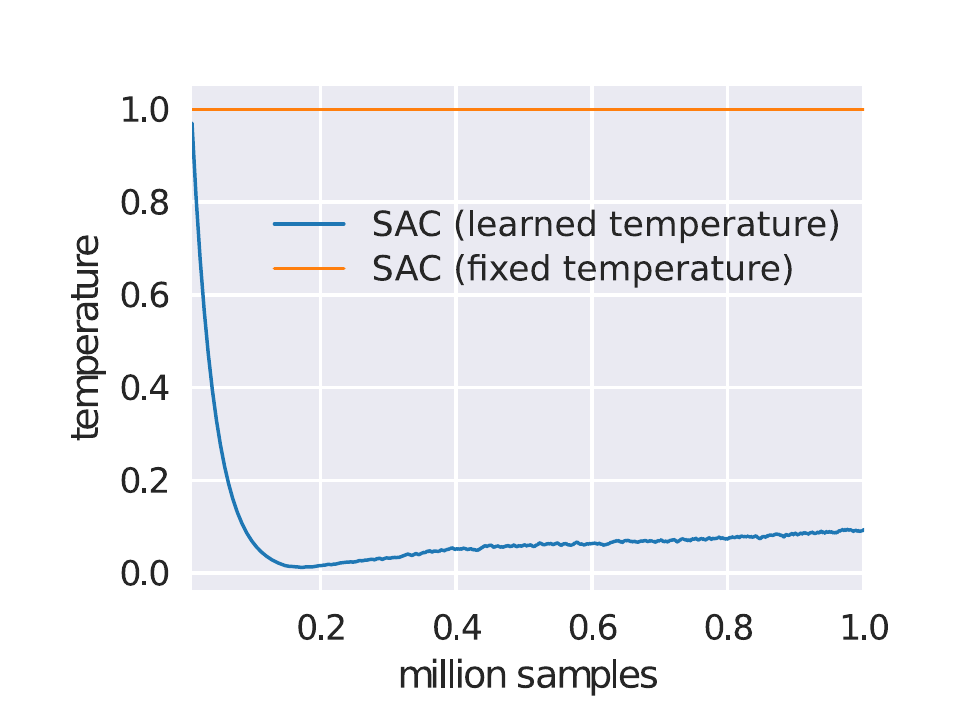}
        \caption{Temperature}
        \label{fig:alpha-half-cheetah}
    \end{subfigure}
    \caption{\small Comparison of our method and standard SAC in terms of entropy and temperature on HalfCheetah. The target entropy for learning the temperature of SAC is -13 in this case.}
\end{figure}

\subsubsection{Validation of Entropy Control}
Next, we compared how the entropy and temperature evolve during training. \autoref{fig:entropy-half-cheetah} compares the  entropy (estimated as an expected negative log probability over a minibatch) on HalfCheetah for SAC with fixed temperature (orange) and our method (blue), which uses a target entropy of -13. The figure clearly indicates that our algorithm is able to match the target entropy in a relatively small number of steps. On the other hand, regular SAC has a fixed temperature parameter and thus the entropy slowly decreases as the Q-function increases.  \autoref{fig:alpha-half-cheetah} compares the temperature parameter of the two methods. Our method (blue) actively adjusts the temperature,  particularly in the beginning of training when the Q-values are small and the entropy term dominates in the objective. The temperature is quickly pulled down so as to make the entropy to match the target. For other simulated environments, we observed similar entropy and temperature curves throughout the learning.

\subsection{Evaluation on Simulated Minitaur Environment}
\label{sec:sim_minitaur_env}
\begin{figure}[tb]
	\begin{subfigure}[b]{0.24\textwidth}
	    \centering
        \includegraphics[width=0.7\textwidth]{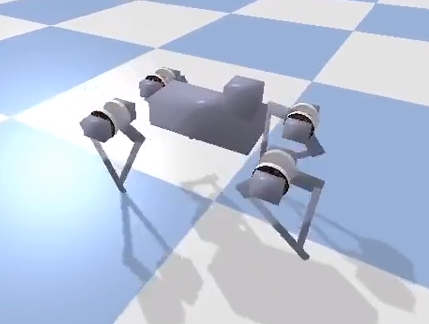}
        \vspace{6mm}
        \caption{Minitaur simulation}
    \end{subfigure}
	\begin{subfigure}[b]{0.24\textwidth}
        \includegraphics[width=\textwidth]{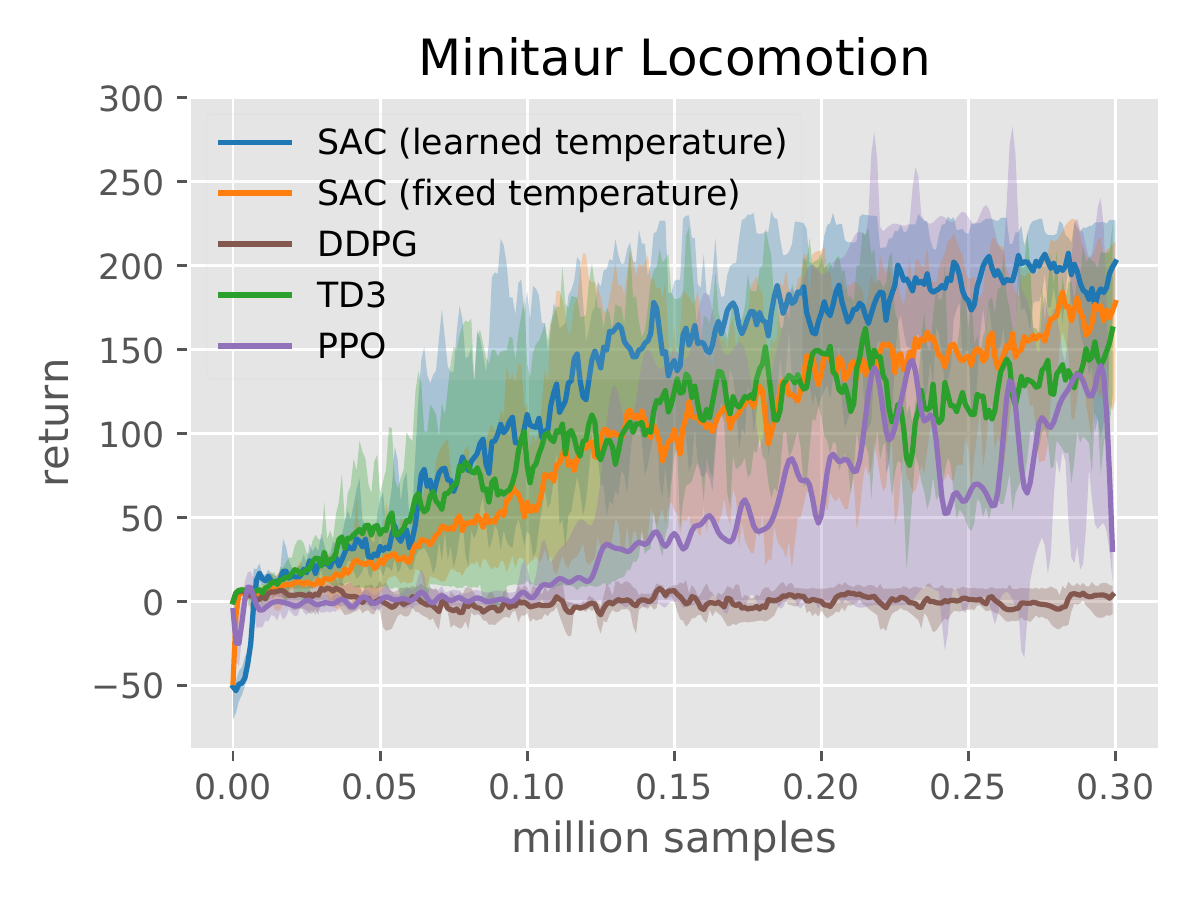}
        \caption{Learning curve}
    \end{subfigure}
	\begin{subfigure}[b]{0.24\textwidth}
        \includegraphics[width=\textwidth]{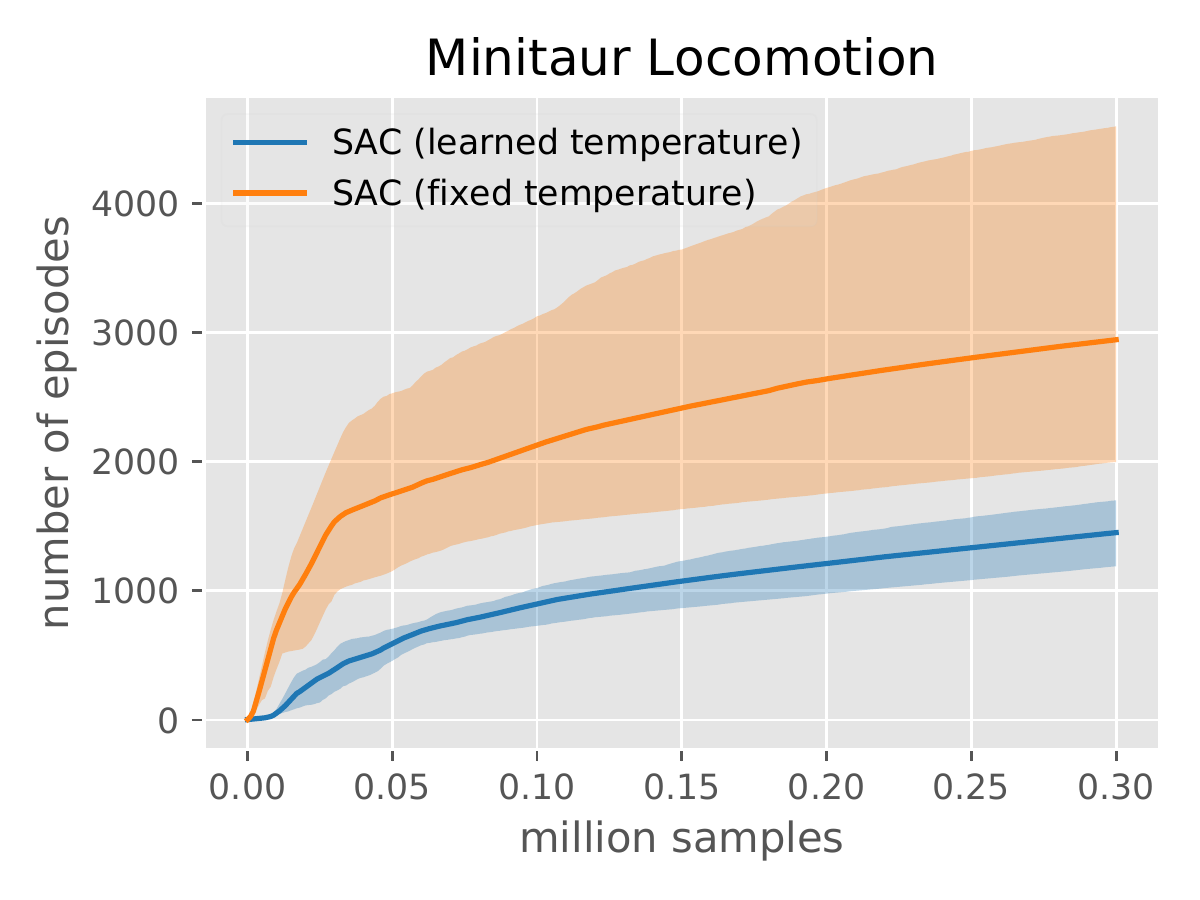}
        \caption{Number of episodes}
    \end{subfigure}
	\begin{subfigure}[b]{0.24\textwidth}
        \includegraphics[width=\textwidth]{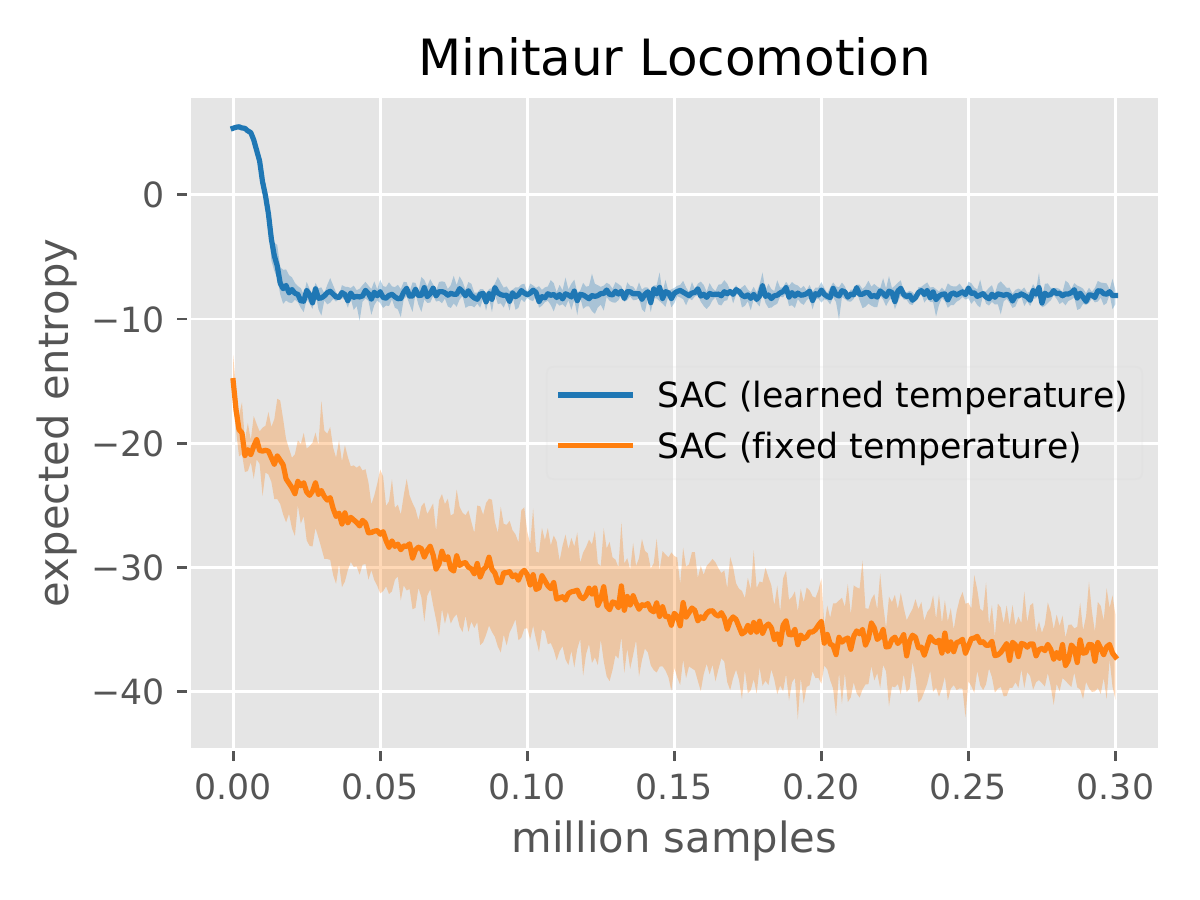}
        \caption{Expected entropy}
    \end{subfigure}
    \caption{\small(a) Illustration of the Minitaur environment. (b) Learning curves. For our method (blue), we used exactly the same hyperparameters as we used for the benchmarks whereas for the baseline (orange), we needed to tune the reward scale. (c) Number of episodes during training. (d) Expected entropy during training.}
    \label{fig:sim_minitaur}
	\vspace{-0.2in}
\end{figure}

Next, we evaluate our method on a simulated Minitaur locomotion task~(\autoref{fig:sim_minitaur}).
Simulation allows us to quantify perturbation robustness, measure states that are not accessible on the robot, and more importantly, gather more data  to evaluate our algorithm. To prevent bias of our conclusion, we have also conducted a careful system identification, following \citet{tan2018sim}, such that our simulated robot moderately represents the real system. However, we emphasize that we do not transfer any simulated policy to the real world ---all real-world experiments use only real-world training, without access to any simulator.

\autoref{fig:sim_minitaur}b compares the learning curve of our method to the state-of-the-art deep reinforcement learning algorithms. Our method is the most data efficient. Note that in order to obtain the result of SAC (fixed temperature) in the plot, we had to sweep though a set of candidate temperatures and choose the best one. This mandatory hyperparameter tuning is equivalent to collecting an order of magnitude more samples, which is not shown in \autoref{fig:sim_minitaur}b. While the \emph{number of steps} is a common measure of data efficiency in the learning community, the \emph{number of episodes} can be another important indicator for robotics because the number of episodes determines the number of experiment reset, which typically is time-consuming and require human intervention. Figure \ref{fig:sim_minitaur}c indicates that our method takes fewer numbers of episodes for training a good policy. In the experiments, our algorithm effectively escapes a local minimum of ``diving forward,'' which is a common cause of falling and early episode termination, by maintaining the policy entropy at higher values (\autoref{fig:sim_minitaur}d). 

The final learned policy in simulation qualitatively resembles the gait learned directly on the robots. We tested its robustness by applying lateral perturbations to its base for $0.5$~seconds with various magnitudes. Even though no perturbation is injected during training, the simulated robot can withstand up to $220$N lateral pushes and subsequently recover to normal walking. This is significantly larger than the maximum $130$N of the best PPO-trained policy that is picked out of $1000$ learning trials. We suspect that this robustness emerges automatically from the SAC method due to entropy maximization at training time.

\section{Learning in the Real World}
\label{sec:minitaur}

In this section, we describe the real-world learning experiments on the Minitaur robot. We aim to answer the following questions:
\begin{enumerate}
    \item Can our method efficiently train a policy on hardware without hyperparameter tuning?
    \item Can the learned policy generalize to unseen situations?
\end{enumerate}

\subsection{Experiment Setup}

Quadrupedal locomotion presents substantial challenges for real-world reinforcement learning. The robot is underactuated, and must therefore delicately balance contact forces on the legs to make forward progress. A suboptimal policy can cause it to lose balance and fall, which will quickly damage the hardware, making sample-efficient learning essential. In this section, we test our learning method and system on a quadrupedal robot in the real world settings. We use the Minitaur robot, a small-scale quadruped with eight direct-drive actuators \cite{kenneally2016design}. Each leg is controlled by two actuators that allow it to move in the sagittal plane. The Minitaur is equipped with motor encoders that measure the motor angles and an IMU that measures the orientation and angular velocity of Minitaur's base. 

In our MDP formulation, the observation includes eight motor angles, roll and pitch angles, and their angular velocities. We choose to exclude the yaw measurement because it drifts quickly. The action space includes the swing angle and the extension of each leg, which are then mapped to desired motor positions and tracked with a PD controller \cite{tan2018sim}. For safety, we choose low PD gains $k_p=0.3$ and $k_d=0.003$ to ensure compliant motion. We find that the latencies in the hardware and the partial observation make the system non-Markovian, which significantly degrades the learning performance. We therefore augment an observation space to include a history of the last five observations and actions which results in a $112$ dimensional observation space. The reward is defined as:
\begin{align*}
r(\st, \at) = &w_1 (x_t - x_{t-1}) - w_2 |{\ddot{\action}}_t| \\
&- w_3 |\phi| - w_4 \sum_{i\in\{1,2\}}{\max(\bar{q} - q_{i}, 0)}.
\label{eq:reward_function}
\end{align*}
The function encourages longer walking distance ($x_t - x_{t-1}$), which is measured using the motion capture system, and penalizes large joint accelerations (${\ddot{\action}}_t$), computed via finite differences using the last three actions. We also find it necessary to penalize a large roll angle of the base ($\phi$) and the joint angles when the front legs ($q_{1}, q_{2}$) are folded under the robot, which are the common failure cases. The weights are set to $1.0, 0.05, 0.5, 1.0$ respectively and the maximum angle threshold $q$ is set to $-0.3$ radians.

We parameterize the policy and the value functions with fully connected feed-forward neural networks with two hidden-layers and 256 neurons per layer, which are randomly initialized. For preventing too jerky motions at the early stage, we smoothed out actions for the first $50$~episodes.

\subsection{Results}
\begin{figure}[tb]
    \centering
    \includegraphics[width=0.30\textwidth]{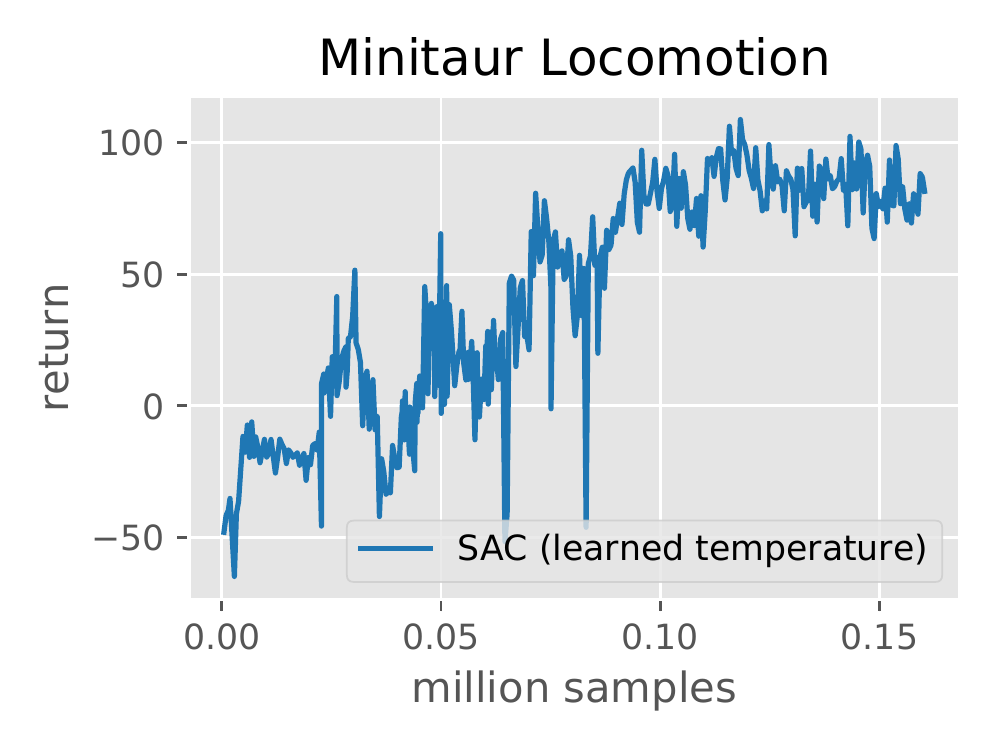}
    \caption{\small Learning curve of SAC with learned temperature on the Minitaur robot.}
	\label{fig:minitaur_real_learning}
\end{figure}

\begin{figure}[tb]
    \centering
	\begin{subfigure}[b]{0.44\textwidth}
	    \centering
	    \hspace{0.5cm}
        \includegraphics[width=0.8\textwidth]{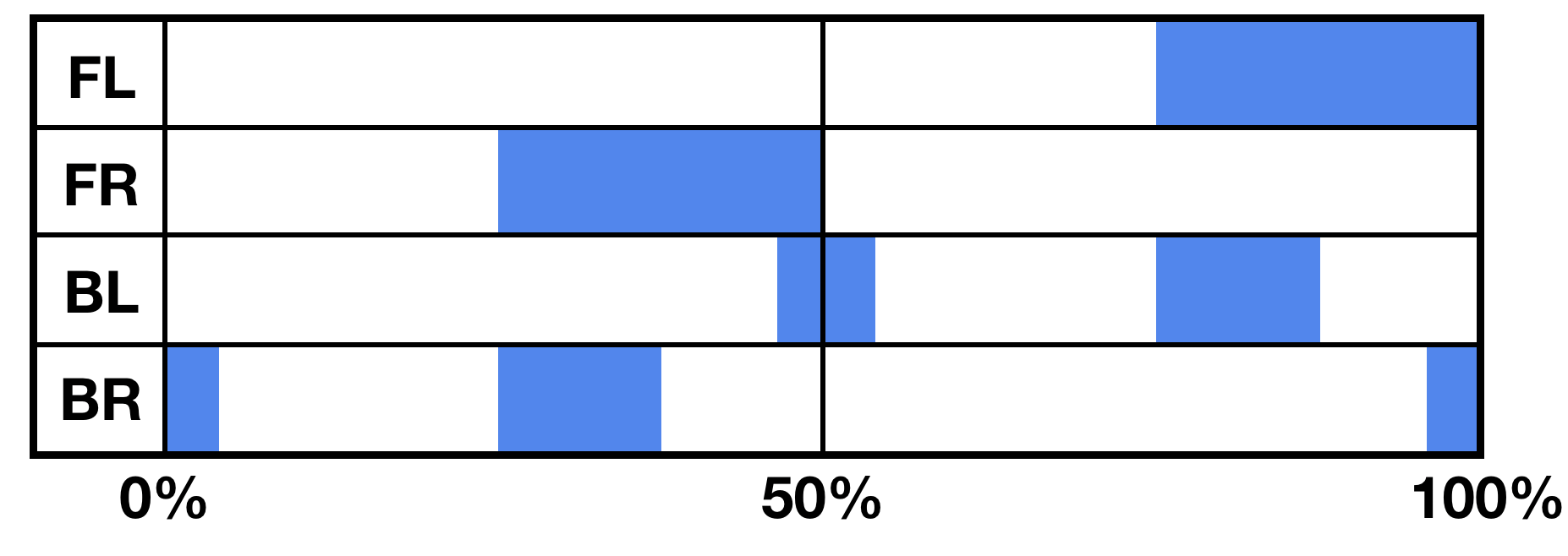}
        \caption{Footfall pattern of learned gait}
    \end{subfigure}
	\begin{subfigure}[b]{0.44\textwidth}
	    \centering
        \includegraphics[width=0.98\textwidth]{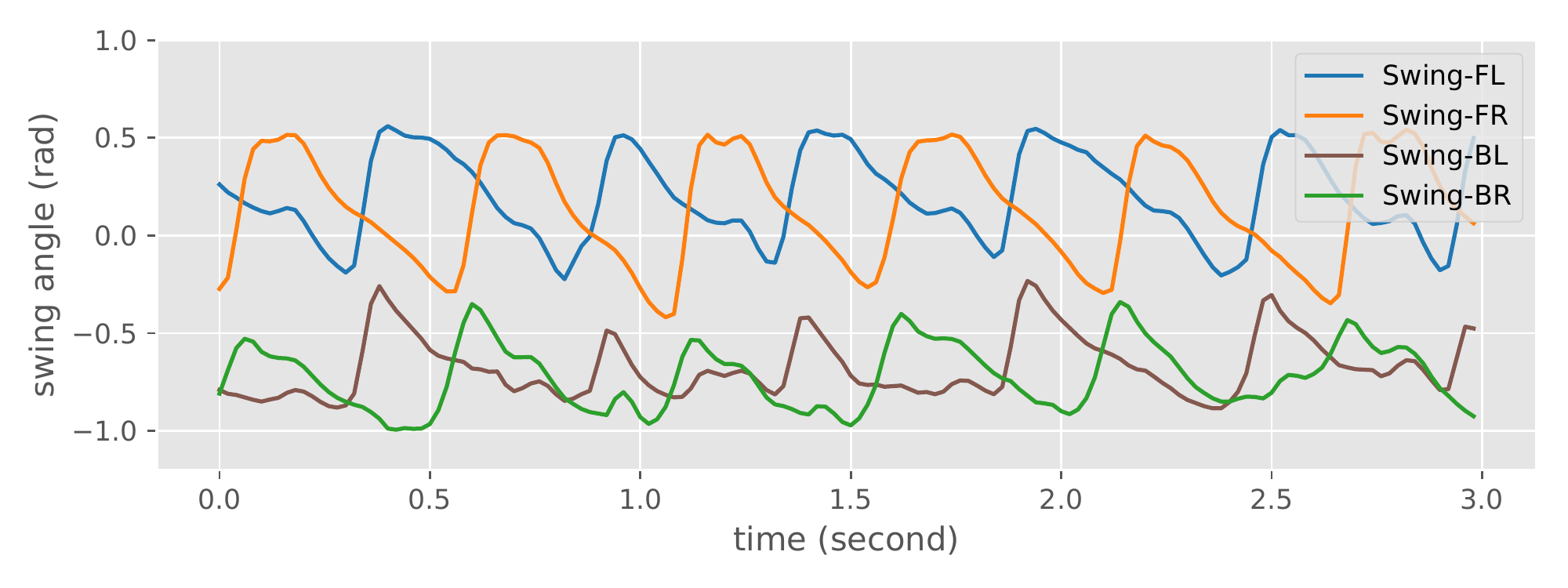}
        \caption{Swing angles of learned gait}
    \end{subfigure}
	\begin{subfigure}[b]{0.44\textwidth}
	    \centering
        \includegraphics[width=0.98\textwidth]{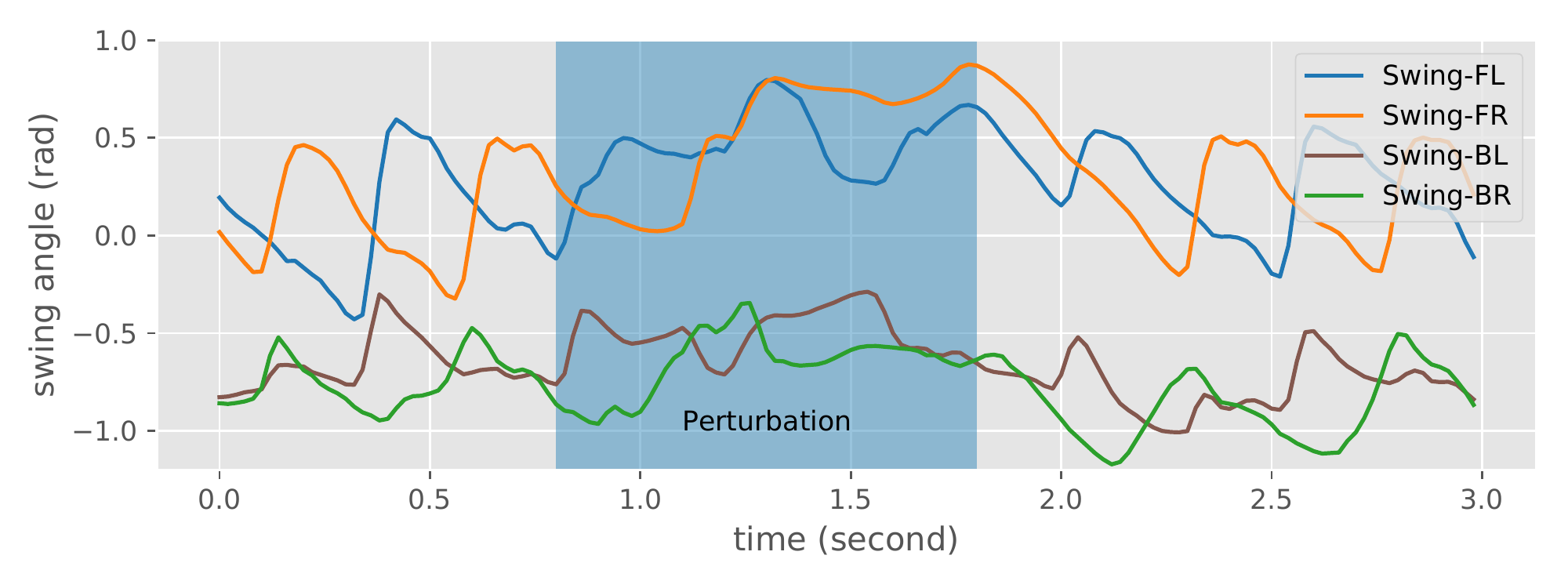}
        \caption{Swing angles of learned gait with perturbations}
    \end{subfigure}    
    \caption{\small Illustration of the learned policy. (a) Footfall pattern of a single cycle of the learned gait. Swing phases are drawn as blue bars. (b) Swing angles of four legs, which are periodic and synchronized. 
    (c) Swing angles with an external perturbation. The learned policy successfully recovers to the nominal gait.}
	\label{fig:joint_trajectory}
\end{figure}

\begin{figure*}[tb]
    \centering
    \begin{subfigure}{0.195\textwidth}
        \includegraphics[width=\textwidth, trim={0 0 0 0}, clip]{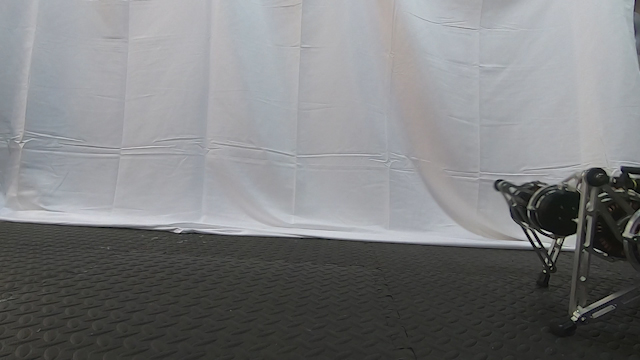}
    \end{subfigure}
    \hfill
    \begin{subfigure}{0.195\textwidth}
        \includegraphics[width=\textwidth, trim={0 0 0 0}, clip]{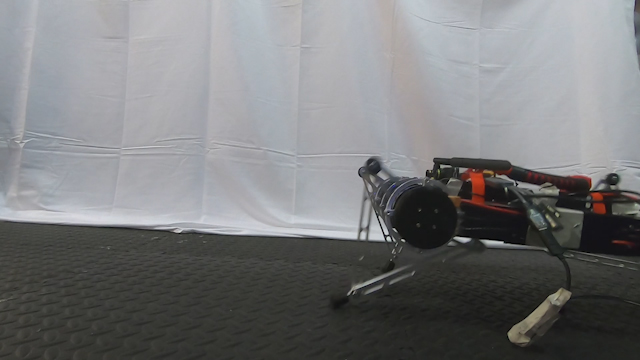}
    \end{subfigure}
    \hfill
    \begin{subfigure}{0.195\textwidth}
        \includegraphics[width=\textwidth, trim={0 0 0 0}, clip]{figures/minitaur/flat/seq15.jpg}
    \end{subfigure}
    \hfill
    \begin{subfigure}{0.195\textwidth}
        \includegraphics[width=\textwidth, trim={0 0 0 0}, clip]{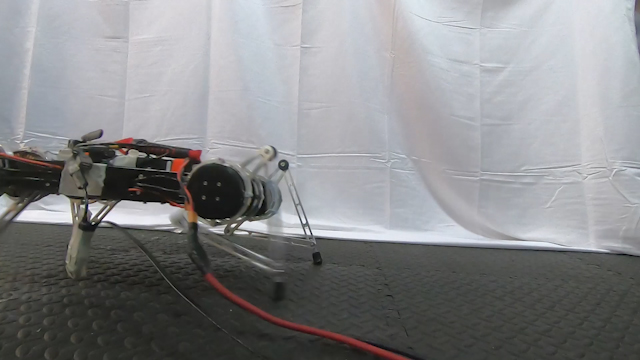}
    \end{subfigure}
    \hfill
    \begin{subfigure}{0.195\textwidth}
        \includegraphics[width=\textwidth, trim={0 0 0 0}, clip]{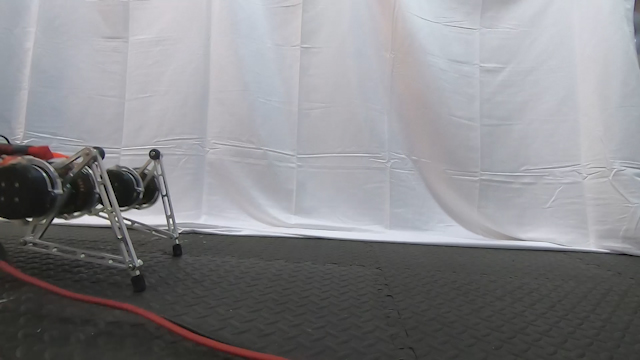}
    \end{subfigure}\\\vspace{1mm}
    
    \begin{subfigure}{0.195\textwidth}
        \includegraphics[width=\textwidth, trim={0 0 0 0}, clip]{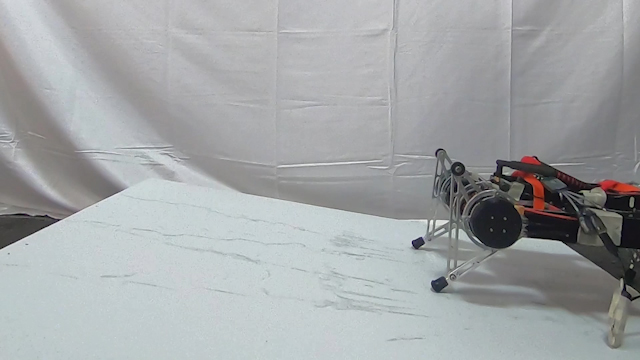}
    \end{subfigure}
    \hfill
    \begin{subfigure}{0.195\textwidth}
        \includegraphics[width=\textwidth, trim={0 0 0 0}, clip]{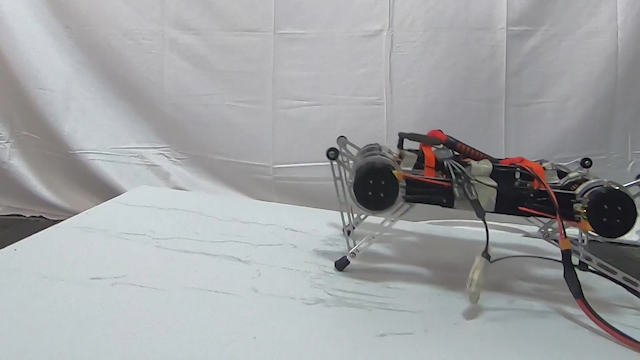}
    \end{subfigure}
    \hfill
    \begin{subfigure}{0.195\textwidth}
        \includegraphics[width=\textwidth, trim={0 0 0 0}, clip]{figures/minitaur/incline/seq20.jpg}
    \end{subfigure}
    \hfill
    \begin{subfigure}{0.195\textwidth}
        \includegraphics[width=\textwidth, trim={0 0 0 0}, clip]{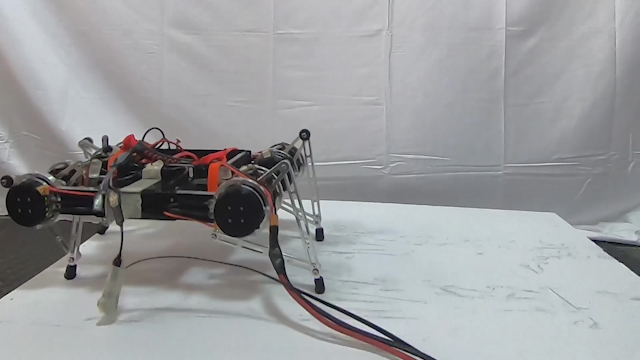}
    \end{subfigure}
    \hfill
    \begin{subfigure}{0.195\textwidth}
        \includegraphics[width=\textwidth, trim={0 0 0 0}, clip]{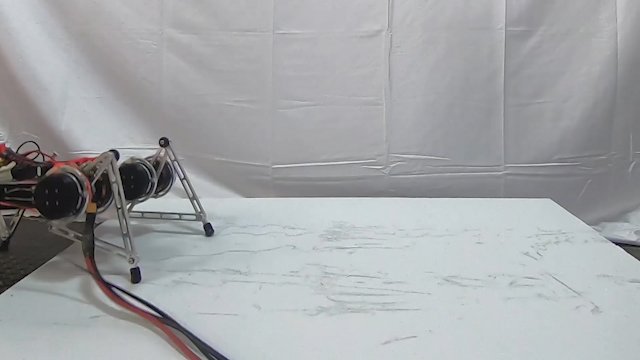}
    \end{subfigure}\\\vspace{1mm}
    
    \begin{subfigure}{0.195\textwidth}
        \includegraphics[width=\textwidth, trim={0 0 0 0}, clip]{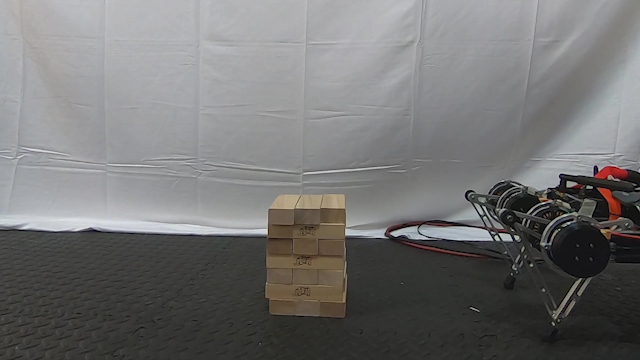}
    \end{subfigure}
    \hfill
    \begin{subfigure}{0.195\textwidth}
        \includegraphics[width=\textwidth, trim={0 0 0 0}, clip]{figures/minitaur/jenga/seq09.jpg}
    \end{subfigure}
    \hfill
    \begin{subfigure}{0.195\textwidth}
        \includegraphics[width=\textwidth, trim={0 0 0 0}, clip]{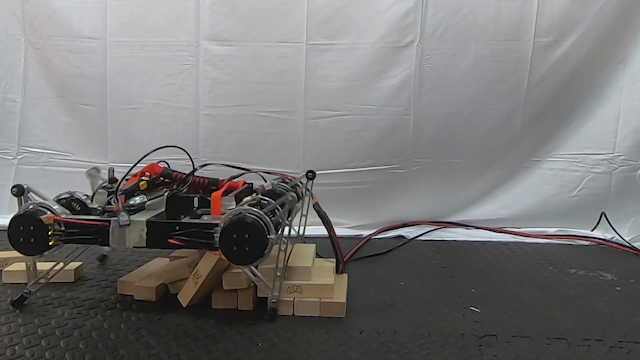}
    \end{subfigure}
    \hfill
    \begin{subfigure}{0.195\textwidth}
        \includegraphics[width=\textwidth, trim={0 0 0 0}, clip]{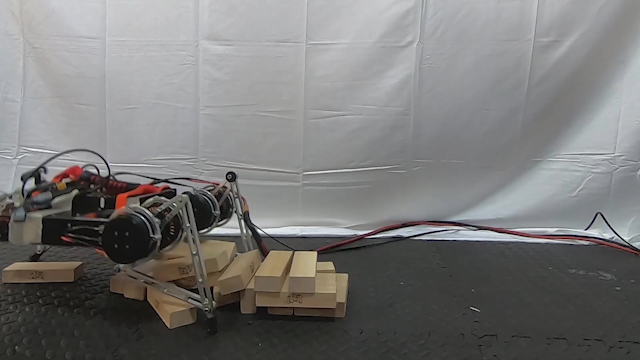}
    \end{subfigure}
    \hfill
    \begin{subfigure}{0.195\textwidth}
        \includegraphics[width=\textwidth, trim={0 0 0 0}, clip]{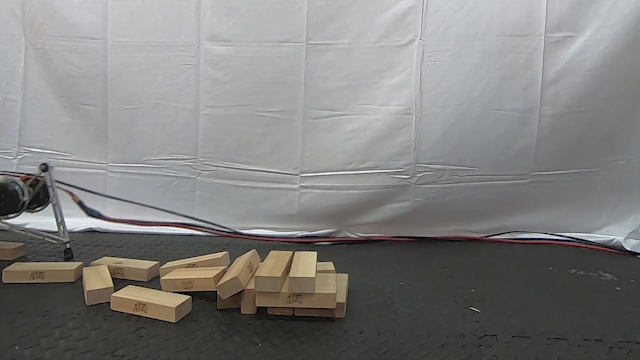}
    \end{subfigure}\\\vspace{1mm}
    
    \begin{subfigure}{0.195\textwidth}
        \includegraphics[width=\textwidth, trim={0 0 0 0}, clip]{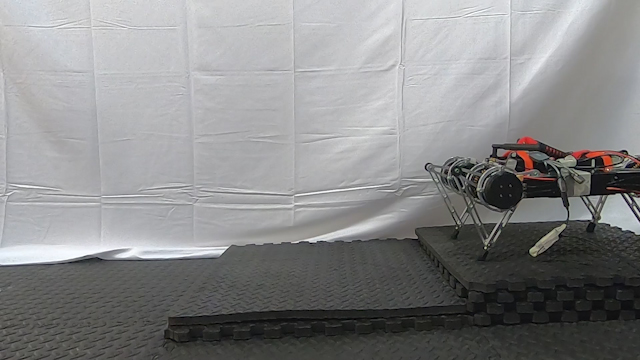}
    \end{subfigure}
    \hfill 
    \begin{subfigure}{0.195\textwidth}
        \includegraphics[width=\textwidth, trim={0 0 0 0}, clip]{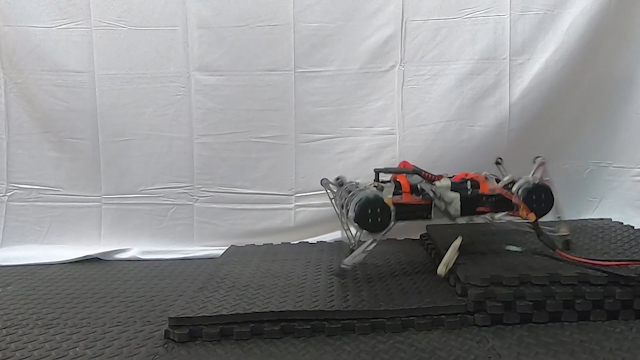}
    \end{subfigure}
    \hfill 
    \begin{subfigure}{0.195\textwidth}
        \includegraphics[width=\textwidth, trim={0 0 0 0}, clip]{figures/minitaur/steps/20181209_demo_step_2_clipped16.jpg}
    \end{subfigure}
    \hfill 
    \begin{subfigure}{0.195\textwidth}
        \includegraphics[width=\textwidth, trim={0 0 0 0}, clip]{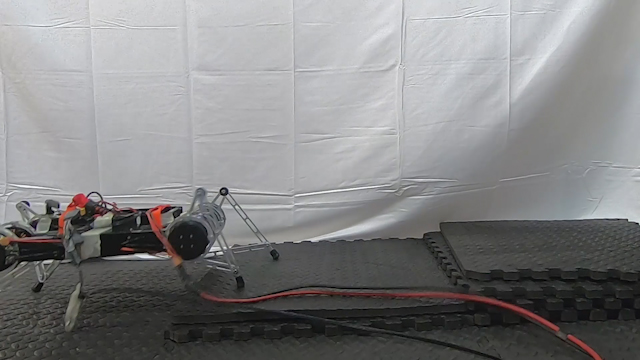}
    \end{subfigure}
    \hfill 
    \begin{subfigure}{0.195\textwidth}
        \includegraphics[width=\textwidth, trim={0 0 0 0}, clip]{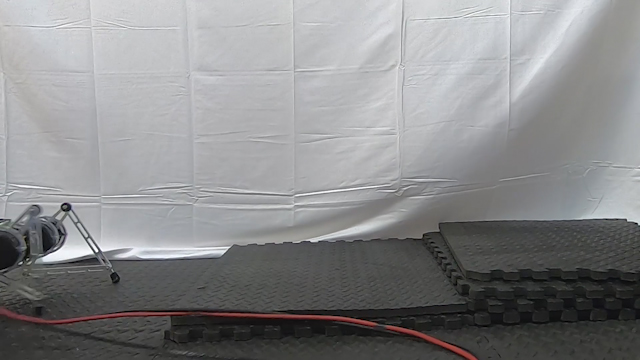}
    \end{subfigure}
    
    \caption{\small We trained the Minitaur robot to walk on flat terrain (first row) in about two hours. At test time, we introduced obstacles, including a slope, wooden blocks, and steps, which were not present at training time, and the learned policy was able to generalize to the unseen situations without difficulty (other rows). 
    }
	\label{fig:minitaur_walking}	
\end{figure*}

Our method successfully learns to walk from 160k control steps, or approximately 400 rollouts. Each rollout has the maximum length of 500 steps (equivalent to 10 seconds) and can terminate early if the robot falls. The whole training process takes about two hours. \autoref{fig:minitaur_real_learning} shows the learning curve,
The performance is slightly less than the simulation, potentially due to fewer collected samples.
Please refer to the supplemental video to see the learning process, the final policy, and more evaluations on different terrains.

The trained robot is able to walk forward at a speed of 0.32m/s ($\sim$0.8 body length per second). The learned gait swings the front legs once per cycle, while pushing against the ground multiple times with the rear legs (\autoref{fig:joint_trajectory}a and b).
Note that the learned gait is periodic and synchronized, though no explicit trajectory generator, symmetry constraint, or periodicity constraint is encoded into the system.
Comparing to the default controller (trotting gait) provided by the manufacturer that walks at a similar speed, the learned gait has similar frequencies ($\sim$2Hz) and swing amplitudes ($\sim$0.7 Rad), but has substantially different joint angle trajectories and foot placement.
The learned gait has a much wider stance and a lower standing height. We evaluated the robustness of the trained policy against external perturbations by pushing the base of the robot backward (\autoref{fig:joint_trajectory}c) for approximately one second, or side for around half second.
Although the policy has never been trained with such perturbations, it successfully recovered and returned to a periodic gait for all $10$ repeated tests.
In the real world, the utility of a locomotion policy hinges critically on its ability to generalize to different terrains and obstacles. Although we trained our policy only on flat terrain (\autoref{fig:minitaur_walking}, first row), we tested it on varied terrains and obstacles (other rows). Because the SAC method learns robust policies due to entropy maximization at training time, the policy can readily generalize to these perturbations without any additional tweaking. The robot is able to walk up and down a slope (second row), ram through an obstacle made of wooden blocks (third row), and step down stairs (fourth row) without difficulty, despite not being trained in these settings. We repeated these tests for $10$ times, and the robot succeeds on all cases.

\section{Conclusion}
We presented a complete end-to-end learning system for locomotion with legged robots. The core algorithm, which is based on a dual formulation of an entropy-constrained reinforcement learning objective, can automatically adjust the temperature hyperparameter during training, resulting in a sample-efficient and stable algorithm with respect to hyperparameter settings. It enables end-to-end learning of quadrupedal locomotion controllers from scratch on a real-world robot. In our experiments, a walking gait emerged automatically in two hours of training without the need of prior knowledge about the locomotion tasks or the robot's dynamic model.
A further discussion of this method and results on other platforms can be found in an extended technical report~\cite{haarnoja2018sac}.
Compared to sim-to-real approaches~\cite{tan2018sim} that require careful system identification, learning directly on hardware can be more practical for systems where acquiring an accurate model is hard and expensive, such as for walking on diverse terrains or manipulation of deformable objects.

To the best of our knowledge, our experiment is the first example of an application of deep reinforcement learning to quadrupedal locomotion directly on a robot in the real world without any pretraining in the simulation, and it is the first step towards a new paradigm of fully autonomous real-world training of robot policies. Two of the most critical remaining challenges of our current system are the heavy dependency on manual resets between episodes and the lack of a safety layer that would enable learning on bigger robots, such as ANYmal~\cite{Hwangboeaau5872} or HyQ\cite{semini2011design}. In the future work, we plan to address these issues by developing a framework for learning policies that are safety aware and can be trained to automatically reset themselves~\cite{EysGuIbaLev18, Hwangboeaau5872}.

\section*{Acknowledgments}
The authors gratefully thank Ken Caluwaerts, Tingnan Zhang, Julian Ibarz,  Vincent Vanhoucke, and the anonymous reviewers for valuable discussion and suggestions.

\bibliographystyle{plainnat}
\bibliography{references}

\end{document}